\documentclass[]{sensenova}

\usepackage{hyperref}
\usepackage{cleveref}
\usepackage{verbatim}

\usepackage{wrapfig}
\usepackage{graphicx}
\usepackage{floatrow}
\usepackage{placeins}
\usepackage{subcaption}
\usepackage{listings}
\usepackage{algorithm}
\usepackage{subcaption}
\usepackage{dsfont}
\usepackage{pifont}

\usepackage{mmstyles}

\usepackage[toc,page,header]{appendix}

\usepackage{xspace}
\usepackage[utf8]{inputenc}  
\usepackage[T1]{fontenc}     
\usepackage{CJKutf8} 

\usepackage[misc]{ifsym}

\usepackage{CJKutf8} 
\usepackage[colorinlistoftodos,prependcaption,textsize=small]{todonotes}

\usepackage{fontawesome5}

\usepackage{soul} 
\definecolor{HLGreen}{rgb}{0.78,0.95,0.78}
\definecolor{HLRed}{rgb}{0.9725,0.8431,0.8549}   
\definecolor{benchblue}{RGB}{58, 122, 252}

\newcommand{\colornum}[1]{%
  \footnotesize
  \ifdim #1 pt < 0pt
    \textcolor{sensepurple}{#1}%
  \else
    \textcolor{benchblue}{#1}%
  \fi
}

\usepackage{cuted} 

\usepackage[table]{xcolor}
\definecolor{bestcolor}{RGB}{219, 208, 237}
\definecolor{secondcolor}{RGB}{241, 237, 248}
\definecolor{thirdcolor}{RGB}{211, 222, 190}
\definecolor{line-blue}{RGB}{243, 248, 252}
\definecolor{line-green}{RGB}{200,242,200}
\definecolor{line-red}{RGB}{255,215,215}

\usepackage{adjustbox}
\usepackage{makecell}
\usepackage{booktabs}

\newlength{\ModelW}
\newlength{\NonModelW}
\newcolumntype{M}{>{\centering\arraybackslash}p{\NonModelW}}
\usepackage{graphicx}
\usepackage{subcaption}

\usepackage{array}      
\usepackage{caption}
\usepackage{subcaption} 

\usepackage{enumitem}

\usepackage{multirow}
\usepackage{makecell}

\usepackage{booktabs,tabularx,array}
\newcolumntype{Y}{>{\centering\arraybackslash}X}
\newcolumntype{L}{>{\raggedright\arraybackslash}X}

\usepackage{framed}
\usepackage{algorithm}
\usepackage{algorithmic}
\usepackage{listings}
\usepackage{soul}          

\definecolor{cotredbg}{RGB}{255,220,220}   
\definecolor{cotgreenbg}{RGB}{220,255,220} 

\title{SenseNova-U1.5: Towards Native Unified Visual Intelligence}

\author{%
\parbox{\textwidth}{\centering
}}

\abstract{

We launch \textbf{SenseNova-U1.5}, an 8B-MoT native unified multimodal model that understands, reasons about, and generates visual content within an encoder-free and VAE-free architecture. 
We strengthen its visual interface through spatially coherent patch reconstruction and scale its training with carefully curated generation and editing data, improved task formulation, structural prompt enhancement, and native resolutions of up to 4K. 
For post-training, we optimize specialized experts for visual aesthetics, bilingual text rendering, infographic generation, and image editing, and consolidate their capabilities through multi-expert on-policy distillation. 
Across extensive evaluations, SenseNova-U1.5 largely advances image fidelity, text rendering, complex composition, multi-reference editing, and interleaved generation, while improving instruction following and preserving subject identity, geometry, and unmodified regions.
Despite limited exposure to structured formats in its generation data, SenseNova-U1.5 generalizes effectively to long, complex, and structured visual instructions, further proving that multimodal understanding can transfer to visual planning and creation. Together, these findings position native unified modelling as a promising path towards systems that perceive, reason and create within a fully end-to-end framework.
We will open-source training code, including supervised fine-tuning, reinforcement learning, and on-policy distillation.

\begin{center}
\includegraphics[width=0.99\textwidth, trim= 180 50 180 20, clip]{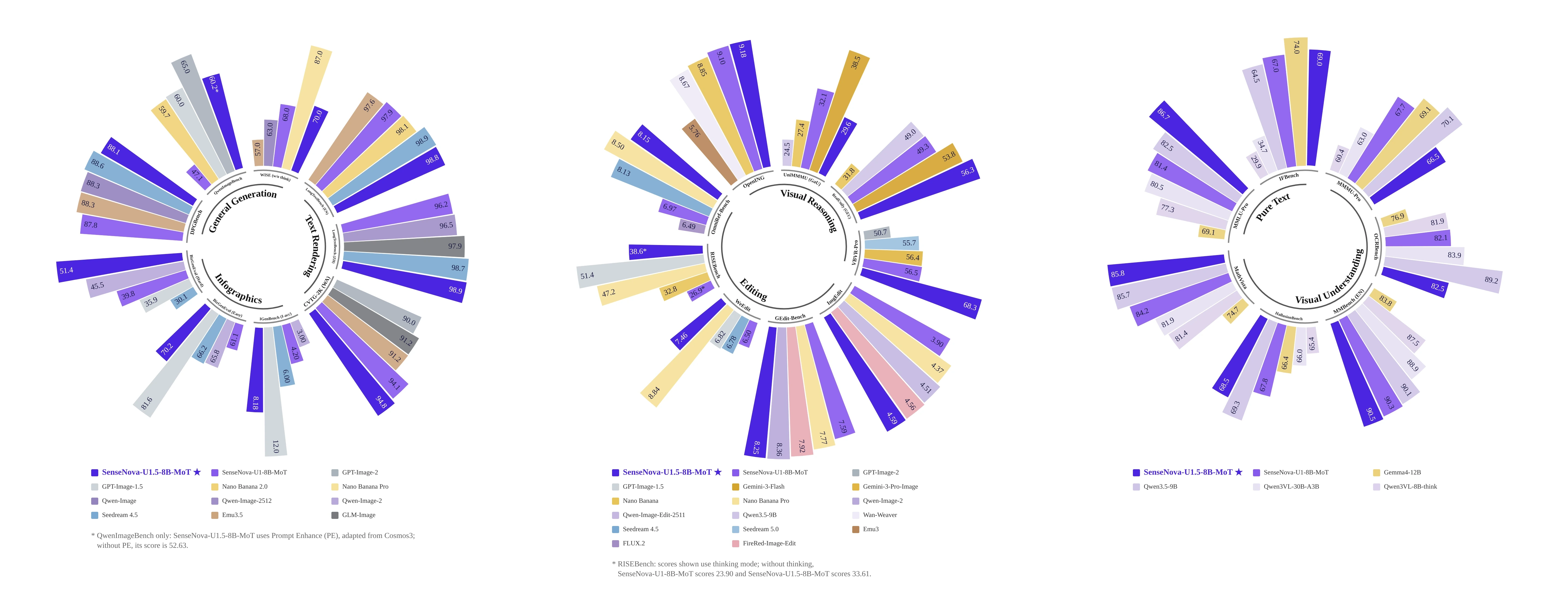}
\end{center}

\checkdata[Official Demo]{\url{https://unify.light-ai.top/}} 
\checkdata[GitHub Code]{\url{https://github.com/OpenSenseNova/SenseNova-U1}}
\checkdata[HuggingFace Model]{\url{https://huggingface.co/collections/sensenova/sensenova-u15}} 
\checkdata[NEO-unify Blog]{\url{https://huggingface.co/blog/sensenova/neo-unify} (March 5, 2026)}
} 
\usepackage{booktabs}
\usepackage{array}
\usepackage{makecell}
\usepackage{multirow}
\usepackage{amsmath}
\usepackage{graphicx}
\usepackage{xcolor}
\usepackage{fontawesome5}
\usepackage{caption}

\providecommand{\frozen}{\textcolor{cyan!70!blue}{\faSnowflake}}
\providecommand{\trainable}{\textcolor{orange!90!red}{\faFire}}

\renewcommand{\arraystretch}{1.10}
\begin{document}
\maketitle

\begin{figure}[H]
  \centering
  \includegraphics[width=\textwidth]{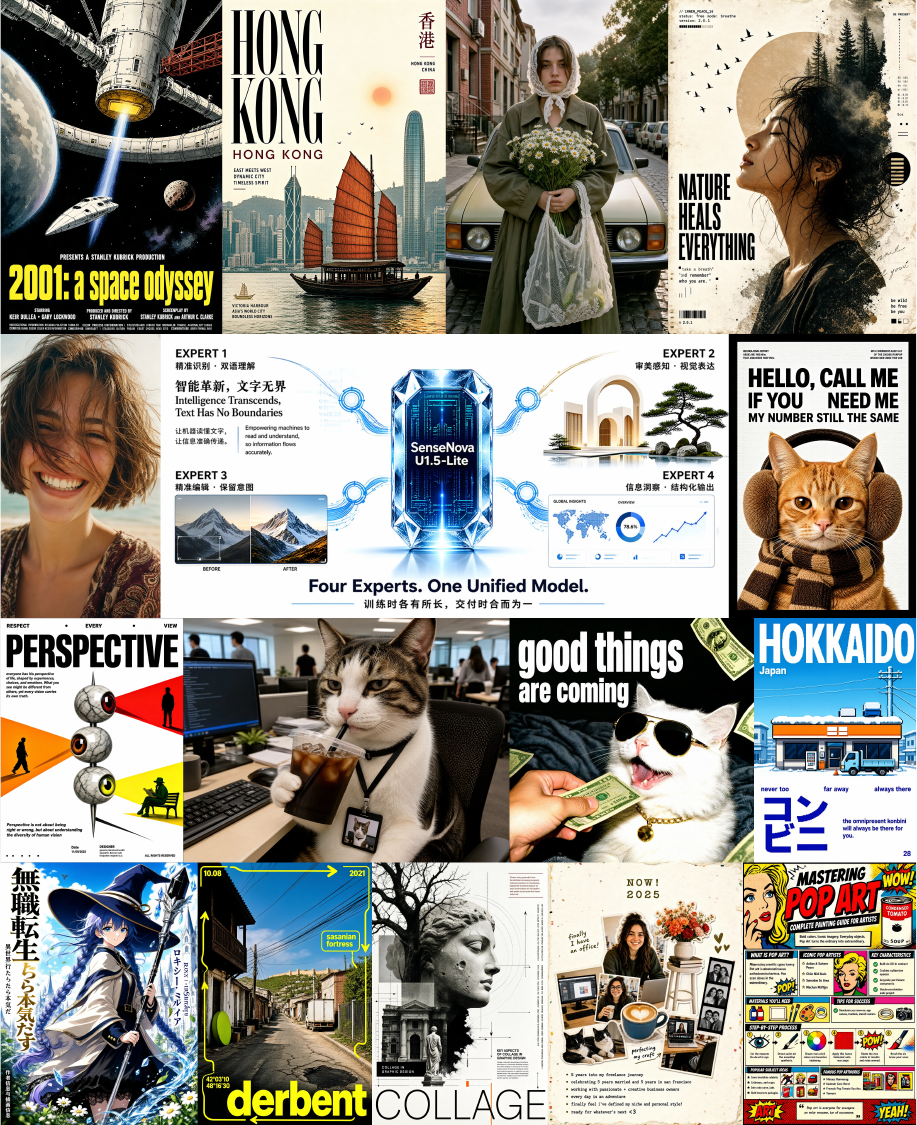}
  \caption{Showcases of SenseNova-U1.5 in infographics and human generation.}
  \label{fig:teaser_figure_t2i}
\end{figure}

\clearpage
\begin{figure}[H]
  \centering
  \includegraphics[width=0.9\textwidth]{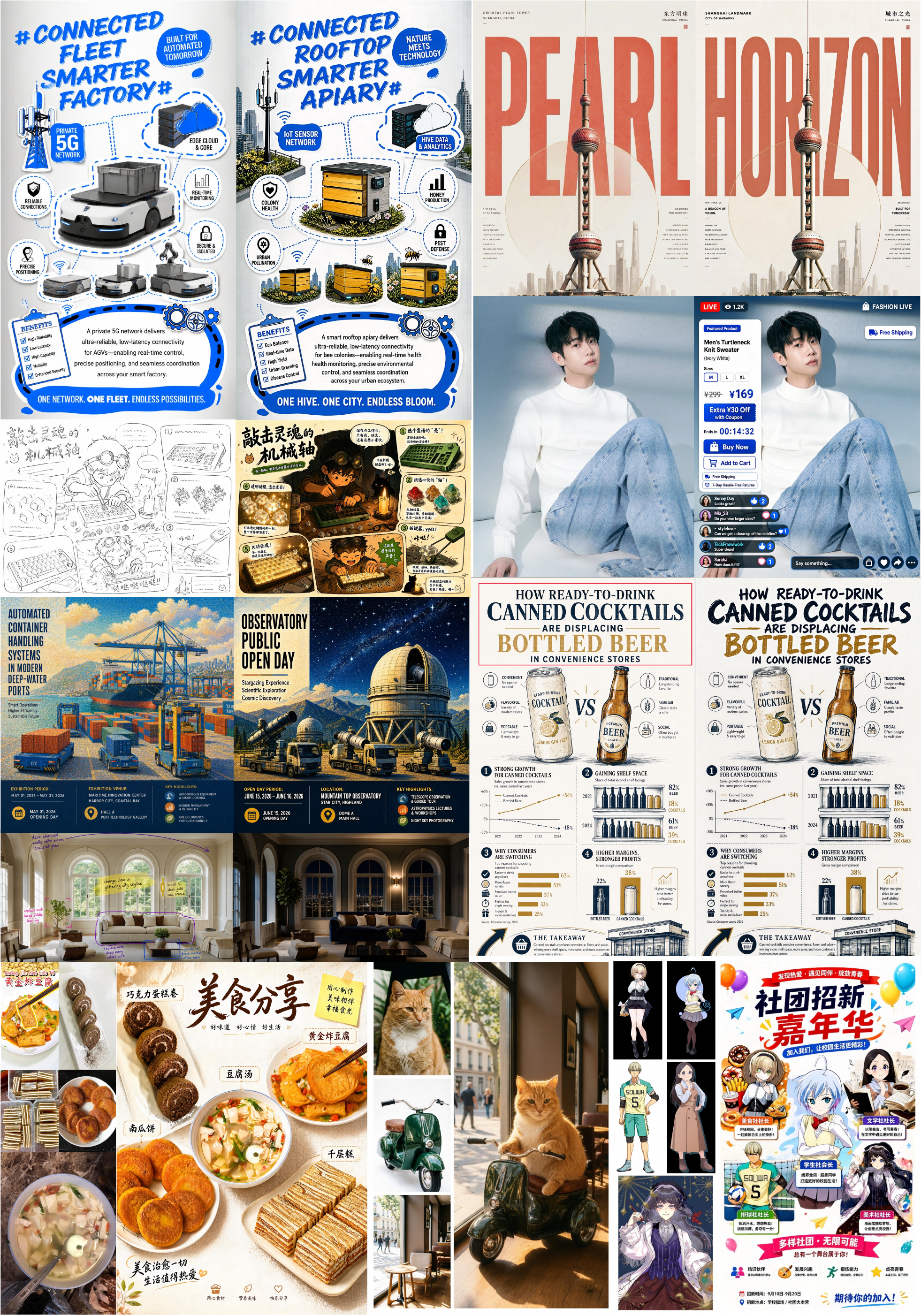}
  \caption{Showcases of SenseNova-U1.5 in image editing and multi-reference generation.}
  \label{fig:teaser_figure_other}
\end{figure}
\clearpage

{
\hypersetup{linkcolor=blue}
\tableofcontents
}
\clearpage

\section{Introduction}
\label{sec:intro}

In recent years, visual generation is rapidly evolving beyond conventional image synthesis into a general medium for visual creation, spanning multilingual typography, information-dense design, high-resolution rendering, multi-reference composition, and fine-grained object-background editing, and interleaved generation. 
However, this broader scope exposes a fundamental architectural divide: most systems perceive images through pretrained vision encoders (VEs)~\cite{sun2023eva,VLP:SigLIP,VLP:CLIP,VLP:SigLIP-2} but generate them through variational autoencoders (VAEs)~\cite{vae,vavae,flux2024,stabilityai2024sd35}. Consequently, understanding and generation operate in distinct representational spaces: one optimized for semantic abstraction, the other for pixel-level fidelity. 
Although effective, this separation limits how seamlessly perception, reasoning, and generation can be coordinated within a unified visual framework across diverse forms of visual creation.

Native unified modelling takes a different route by learning directly from pixels and words. SenseNova-U1~\cite{diao2026sensenovau1} established the feasibility of this paradigm through the encoder-free and VAE-free NEO-unify~\cite{sensenova2026neounify} architecture, bringing perception, reasoning, and pixel-space generation into a single end-to-end model. Its deliberately lightweight visual interface, however, reconstructs each visual token as an independent RGB patch. Although efficient, this factorization prevents information exchange across neighbouring patches during the final stage of image formation, making seams, texture discontinuities, and geometric inconsistencies increasingly pronounced at high resolutions.

Here we present \textbf{SenseNova-U1.5}, an 8B-MoT native unified multimodal model for visual understanding, reasoning, and generation. At its core is a first major architectural shift: from \textit{independent patch prediction} to \textit{spatially joint reconstruction}. Rather than mapping each visual token to pixels in isolation, we project tokens onto a two-dimensional feature field and progressively reconstruct the image through spatial convolutions and Pixel Shuffle upsampling. This allows neighbouring regions to exchange information before pixels are finalized, so colour, texture, and geometry can be resolved jointly rather than patch by patch. The resulting design preserves the efficiency of compact visual sequences while substantially improving spatial coherence and enabling native generation at resolutions of up to 4K.

The second major advance is a shift from \textit{joint reward optimization} to a \textit{specialize-then-unify strategy}. Visual creation spans fundamentally different capabilities, from aesthetic synthesis and bilingual text rendering to infographic design and image editing, each with distinct reward signals, rollout dynamics, and optimization challenges. Jointly optimizing them within a single policy can entangle competing objectives and dilute task-specific gains. 
We therefore first specialize and then unify: dedicated reinforcement-learning (RL) experts are optimized for distinct capabilities with tailored data, rewards, sampling strategies, and regularization, and their complementary strengths are subsequently consolidated through multi-expert on-policy distillation. By distilling along the student model’s own generation trajectories, this process transfers expert capabilities into a single unified policy while preserving their complementary strengths.

Across extensive evaluations, SenseNova-U1.5 further demonstrates that a single native visual representation can serve as a shared substrate for understanding, reasoning, generation, and editing, eliminating both parallel visual pathways and repeated conversion between encoder features and VAE latents. Despite compressing each \(32\times32\)-pixel region into a single visual token, this compact representation supports efficient inference while delivering strong performance across image fidelity, bilingual typography, complex composition, multi-reference editing, interleaved generation, instruction following, visual preservation, and fine-grained control.
More strikingly, SenseNova-U1.5 generalizes to long, compositional, and highly structured visual instructions despite limited reliance on fixed generation templates during training. This suggests that structural knowledge and planning capabilities acquired through multimodal understanding and reasoning can transfer naturally to visual creation. 
Together, these results show that native unification is more than an architectural simplification: a compact shared visual representation can efficiently support both seeing and creating, while enabling capabilities learned through one form of visual intelligence to reinforce another.

\section{Related Works}
\label{sec:related_works}

\subsection{Native Multimodal Unified Models}

Native vision-language models (VLMs), including Fuyu-8B~\cite{VLM:Fuyu-8b}, EVE~\cite{VLM:EVE,VLM:EVEv2}, Mono-InternVL~\cite{VLM:Mono-InternVL,VLM:Mono-InternVL-1.5}, NEO~\cite{Diao2025NEO,diao2026neov}, Gemma4-12B~\cite{gemma42026}, and Inkling~\cite{inkling}, directly process visual inputs without external encoders and have progressively narrowed the performance gap with leading modular VLMs~\cite{wang2025internvl3,qwen35blog,openai_gpt5_systemcard,gemini_3_pro_systemcard}. 
A parallel trend is emerging in visual generation, where recent works directly model pixels~\cite{PixelFlow, DiP, yu2025pixeldit,li2025back}, demonstrating that high-fidelity synthesis need not rely on heavily compressed latent spaces~\cite{vae,vqvae}.
Building on these advances, native unified multimodal models increasingly seek to combine understanding and generation within a single framework. 
Among them, discrete native models~\cite{Chameleon,MOMA,wang2024emu3,cui2025emu35nativemultimodalmodels,MoT,li2025onecat,team2026longcat} unify multimodal learning through token-level autoregression, while continuous native approaches~\cite{sensenova2026neounify,diao2026sensenovau1,tuna2,cai2026hidream} pursue end-to-end modeling without explicit tokenizers or latent bottlenecks. Building on NEO-unify~\cite{sensenova2026neounify}, our SenseNova-U series further scales this direction toward a fully native foundation where understanding, reasoning, and generation emerge from a shared visual substrate.

\subsection{Reinforcement Learning for Diffusion Models}

In language modeling, reinforcement learning from human feedback (RLHF) established a general post-training paradigm where learned rewards guide policy optimization, with PPO stabilizing updates against a reference policy~\cite{ouyang2022training,schulman2017proximal}. Notably, GRPO and DAPO improve efficiency and scalability by removing explicit value models and introducing group-relative advantages and enhanced online optimization strategies~\cite{shao2024deepseekmath,yu2025dapo}.
Online RL has also been extended to visual generation, with ReFL, DDPO, and DPOK optimizing diffusion models via preference rewards or policy gradients~\cite{xu2023imagereward,black2023ddpo,fan2023dpok}, while AlignProp and D3PO improve efficiency and reduce reliance on explicit reward models~\cite{prabhudesai2023alignprop,yang2024d3po}.
More recently, Flow-GRPO and DanceGRPO extend group-relative optimization to diffusion and flow-matching models, improving preference alignment, compositional accuracy, text rendering, and visual quality~\cite{liu2025flowgrpo,xue2025dancegrpo}.

For flow-matching models, online RL requires stochastic exploration beyond deterministic ordinary differential equation (ODE) inference. Flow-GRPO~\cite{liu2025flow} enables such exploration through stochastic differential equation (SDE)-based rollouts, while coefficients-preserving sampling (CPS)~\cite{wang2025cps} and Precise~\cite{zou2026precise} improve finite-step sampling by better preserving the underlying flow dynamics and balancing exploration with distributional fidelity. Beyond sampling, GRPO-Guard~\cite{wang2025grpoguard} mitigates reward over-optimization through regulated clipping and noise-aware gradient reweighting. Existing visual RL further relies on capability-specific rewards, including preference models for perceptual quality and text--image alignment~\cite{xu2023imagereward,wu2023hps,ma2025hpsv3,liu2026hpsv3pp}, as well as specialized rewards for typography and editing~\cite{huang2025customizedgrpo,guo2026editr1}. Motivated by these advances, we adopt task-dependent RL post-training that combines CPS or Precise sampling with interleaved or task-specific rewards to address the distinct optimization requirements of generation and editing.

\subsection{On-Policy Distillation for Unified Models}

Recently, on-policy distillation (OPD)~\cite{agarwal2024onpolicy,lu2025onpolicydistillation} mitigates the distribution mismatch of conventional knowledge distillation by training the student on its own generated trajectories while receiving dense supervision from a teacher. Building on this paradigm, MOPD~\cite{ma2026mopd} extends OPD to multi-teacher capability integration, where independently optimized domain experts supervise the student's on-policy rollouts, enabling diverse reasoning capabilities to be consolidated into a single model. 
Beyond language modeling, OPD has been explored for multimodal understanding and reasoning~\cite{bousselham2026vold,yuan2026vision}, where teachers supervise student-generated multimodal trajectories. Its application has also recently expanded to visual generation.
Specifically, Flow-OPD, DiffusionOPD, and DanceOPD distill task-specialized generators along student-generated denoising trajectories through teacher-consistency signals, transition matching, and velocity-field regression, respectively~\cite{fang2026flow,li2026diffusionopd,zhou2026danceopd}. Besides, DiffusionOPSD instead removes the external teacher and constructs bounded self-distillation targets from differentiable reward gradients~\cite{zhou2026diffusionopsd}. Our setting is complementary: rather than forcing all tasks into a shared distillation recipe, we retain four task-specialized external experts and hard-route their supervision into a single native pixel-space model. This design preserves the task-dependent conditioning, guidance, and resolution policies of each expert, while consolidating their specialized capabilities into a unified model.
\section{Methodology}
\label{sec:methodology}
\subsection{Model architecture}

\begin{figure}[t]
  \centering
  \includegraphics[width=0.99\textwidth]{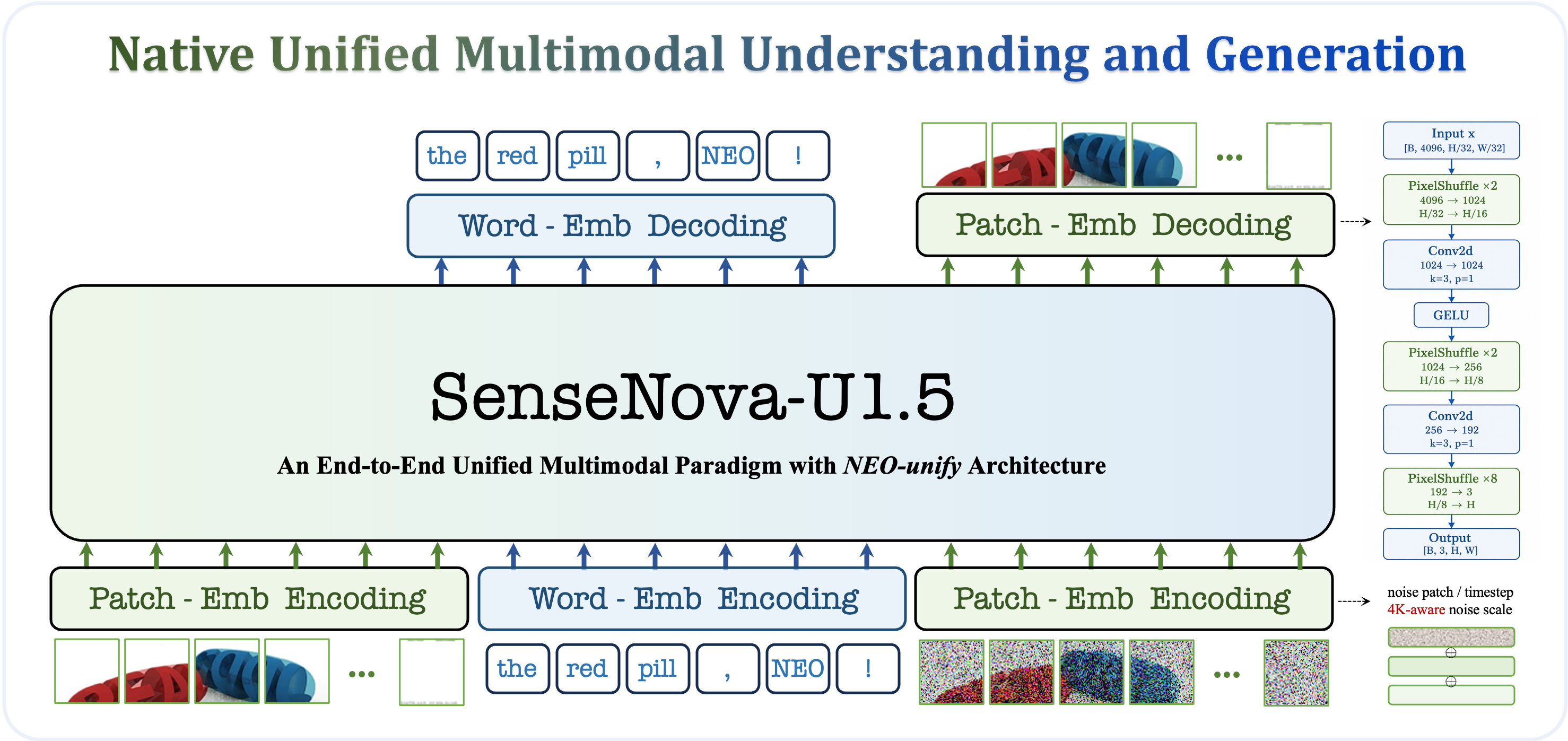}
  \caption{\textbf{Overview of SenseNova-U1.5.} Compared with SenseNova-U1, U1.5 further refines the near-lossless visual interface on both encoding and decoding: resolution-aware noise conditioning is extended to the \(4096\times4096\) range, while the original patch-wise MLP head is replaced by a lightweight spatial decoder with Pixel Shuffle and \(3\times3\) convolutions. These improvements preserve compact \(32\times32\) visual tokenization while enhancing spatial continuity, high-resolution fidelity, and downstream robustness.
 }
  \label{fig:u15_arch}
\end{figure}

\textbf{Near-Lossless Visual Interface.}
SenseNova-U1.5 retains the lightweight native visual interface introduced in NEO~\cite{Diao2025NEO}, directly transforming raw images or noise-corrupted visual inputs into compact token sequences without relying on an external visual encoder or VAE.
Specifically, two convolutional projections with GELU activations downsample the input by factors of \(16\) and \(2\), respectively, yielding one visual token for each \(32\times32\) image region.
Two-dimensional sinusoidal positional embeddings preserve spatial coordinates, while special \texttt{<img>} and \texttt{</img>} tokens delimit individual visual blocks.
Text is tokenized with the original language tokenizer, after which visual and textual representations are projected into a shared hidden space and jointly processed by the unified backbone.
This design preserves near-lossless visual information while keeping the sequence length tractable for large-scale multimodal modeling.

For generation, the effective noise magnitude varies with image resolution, making resolution an important conditioning signal for the denoising process.
SenseNova-U1.5 therefore explicitly incorporates a resolution-dependent noise-scale embedding.
Compared with SenseNova-U1, whose normalization range is calibrated up to \(2048\times2048\), we extend the reference resolution to \(4096\times4096\) to better support native high-resolution synthesis in Figure~\ref{fig:u15_arch}.
Let \(\sigma_R(H,W)\) denote the noise scale associated with resolution \((H,W)\).
We normalize it as
$
\bar{\sigma}_R
=
{\sigma_R(H,W)}/\sigma_{\max}
$, where \(\sigma_{\max}\) corresponds to the \(4096\times4096\) reference resolution, and encode it using a dedicated sinusoidal MLP, \(\mathrm{NSEmb}(\cdot)\).
The resulting embedding is combined with the diffusion timestep representation:
$
\mathbf{s}_t
=
\boldsymbol{\tau}_t
+
\mathrm{NSEmb}\!\left(\bar{\sigma}_R\right)
$, providing the denoiser with explicit awareness of both diffusion time and resolution-dependent noise statistics.
This conditioning helps the model adapt more consistently to variations across diverse image resolutions and aspect ratios.

\begin{table*}[t]
\centering
\small
\setlength{\tabcolsep}{5pt}
\renewcommand{\arraystretch}{1.15}
\resizebox{\textwidth}{!}{
\begin{tabular}{l|ccccccc}
\specialrule{1.2pt}{0pt}{0pt}
\textbf{Configuration}
& \textbf{Patch Size}
& \textbf{Pre-Buffer}
& \textbf{\# Layers}
& \textbf{\# Heads (Q/KV)}
& \textbf{Head Size (T/H/W)}
& \textbf{Hidden Size}
& \textbf{\# Parameters (Und/Gen)} \\
\midrule
\textbf{SenseNova-U1.5}
& $32 \times 32$
& \ding{51}
& 42
& 32 / 8
& 64 / 32 / 32
& 4,096
& 8.2B / 8.2B \\
\specialrule{1.2pt}{0pt}{0pt}
\end{tabular}
}
\caption{\textbf{Configurations of SenseNova-U1.5.} It features a high spatial compression ratio, Pre-Buffer design, native RoPE for unified spatiotemporal encoding, and a Mixture-of-Transformers architecture for joint multimodal understanding and generation.}
\label{tab:model_config}
\end{table*}

Compact patchification greatly reduces image-modeling cost, but independently decoding each \(32\times32\) token with an MLP imposes an undesirable patch-wise factorization on the output. This is particularly problematic at high resolutions, where local image continuity makes independent patch prediction prone to seams, grid artifacts, and texture discontinuities. SenseNova-U1.5 therefore replaces the original MLP head with a lightweight spatially coupled decoder.
Given backbone hidden states
$\mathbf{H}
\in
\mathbb{R}^{B\times hw\times d}
$, we first restore their two-dimensional topology,
$
\mathbf{H}_{\mathrm{2D}}
\in
\mathbb{R}^{B\times d\times h\times w}
$, and progressively reconstruct the full-resolution RGB image through Pixel Shuffle stages with upsampling factors of \(2\), \(2\), and \(8\) in Figure~\ref{fig:u15_arch}.
Between successive upsampling stages, \(3\times3\) convolutions enable information exchange across neighboring token regions, allowing pixels near patch boundaries to be determined jointly rather than independently.
The decoder complements compact token-space modeling by restoring local spatial interactions before pixel synthesis, enabling global semantics, composition, and fine-grained structure to be jointly optimized within the same end-to-end framework. Trained together with the backbone under the flow-matching objective, this lightweight design improves cross-patch consistency and local continuity with only modest additional computation, substantially reducing boundary artifacts while enhancing the stability of high-resolution generation and downstream adaptation.

\textbf{Native Mixture-of-Transformers.}
SenseNova-U1.5 retains the native Mixture-of-Transformers (MoT) design~\cite{diao2026sensenovau1}, integrating understanding and generation within a single Transformer backbone rather than completely separating them into two independent networks. Clean image-text context and noise-conditioned visual states are interleaved within a unified sequence, allowing semantic representations, visual evidence, and generative dynamics to interact directly through shared self-attention. This design lets generation draw continuously on representations formed during multimodal understanding, without introducing auxiliary fusion modules or cross-space feature conversion.

The attention pattern is structured to reconcile causal language modeling with bidirectional visual interaction. Text tokens attend only to preceding context, while tokens within each clean image block attend bidirectionally to capture spatial dependencies. Noise-conditioned generation tokens likewise interact bidirectionally within their image block and attend to all preceding clean multimodal context. The reverse path is explicitly masked, preventing clean representations from accessing stochastic generation states. This asymmetric information flow exposes generation to rich semantic and visual context while preserving the integrity of representations used for understanding and reasoning.

Crucially, architectural unification does not imply parameter sharing everywhere. Understanding and generation retain separate attention projections, normalization layers, and feedforward modules, dynamically routed by token type at each Transformer layer. Shared attention therefore serves as the interface for cross-stream communication, while stream-specific parameters preserve the distinct computation required by perception and synthesis. This combination of dense interaction and parameter specialization allows the two capabilities to benefit from a common representational space without forcing their different optimization objectives into the same computational pathway.

\noindent\textbf{Unified Training Objectives.}
SenseNova-U1.5 is trained with a unified objective that jointly couples autoregressive language modeling, native pixel-space flow learning, and perceptual supervision.
For multimodal understanding, we optimize the conditional likelihood of the target text sequence as follows:
\begin{equation}
\mathcal{L}_{\mathrm{AR}}
=
-\frac{1}{N}
\sum_{n=1}^{N}
\log p_{\theta}
\left(
x_n \mid x_{<n}, \mathbf{c}
\right),
\end{equation}
where \(x_n\) is the \(n\)-th target token and \(\mathbf{c}\) denotes the preceding multimodal context.

For visual generation, SenseNova-U1.5 follows the \(\mathbf{x}\)-prediction~\cite{li2025back} and learns pixel-space flow matching directly in RGB space. To accommodate varying image resolutions, we construct the training trajectory with resolution-adaptive noise, sampling \(\boldsymbol{\epsilon}\sim\mathcal{N}(0,\mathbf{I})\) and \(t\in[0,1]\) to form
\begin{equation}
\mathbf{z}_t
=
t\mathbf{x}
+
(1-t)\sigma_R(H,W)\boldsymbol{\epsilon},
\end{equation}
where \(\sigma_R(H,W)\) adjusts the noise magnitude according to the target resolution.
After that, the model attempts to predict the clean endpoint \(\hat{\mathbf{x}}_{\theta}\), which induces the velocity estimate:
\begin{equation}
\mathbf{v}_{\theta}
=
\frac{\hat{\mathbf{x}}_{\theta}-\mathbf{z}_t}{1-t},
\qquad
\mathbf{v}^{\star}
=
\frac{\mathbf{x}-\mathbf{z}_t}{1-t}.
\end{equation}
Finally, we supervise this trajectory with the pixel-space flow-matching loss as follows: 
\begin{equation}
\mathcal{L}_{\mathrm{Flow}}
=
\mathbb{E}
\left[
\left\|
\mathbf{v}_{\theta}
-
\mathbf{v}^{\star}
\right\|_2^2
\right].
\end{equation}

Beyond pixel-space trajectory supervision, SenseNova-U1.5 further introduces a perceptual objective to improve structural consistency and local visual coherence:
\begin{equation}
\mathcal{L}_{\mathrm{Perc}}
=
\mathrm{LPIPS}
\left(
\hat{\mathbf{x}}_{\theta},
\mathbf{x}
\right),
\end{equation}
where LPIPS~\cite{zhang2018unreasonableeffectivenessdeepfeatures} provides feature-space supervision complementary to the RGB-space flow objective.
The entire training process is jointly optimized under a unified objective:
\begin{equation}
\mathcal{L}
=
\lambda_{\mathrm{AR}}\mathcal{L}_{\mathrm{AR}}
+
\lambda_{\mathrm{Flow}}\mathcal{L}_{\mathrm{Flow}}
+
\lambda_{\mathrm{Perc}}\mathcal{L}_{\mathrm{Perc}}.
\label{eq:total_loss}
\end{equation}
This brings three complementary signals into a single native model: autoregressive supervision establishes semantic understanding and multimodal reasoning, flow matching learns the continuous transformation from noise to images directly in pixel space, and perceptual supervision further regularizes the generated endpoint toward coherent global structure, faithful local textures, and visually plausible appearance. Together, they couple high-level semantics with low-level visual formation, enabling understanding and generation to be learned within a unified representation.

\begin{table*}[!tp]
\centering
\captionsetup{skip=3pt,width=\textwidth}
\begingroup
\footnotesize
\setlength{\tabcolsep}{4pt}
\renewcommand{\arraystretch}{1.0}

\begin{tabular*}{\textwidth}{@{\extracolsep{\fill}}>{\raggedright\arraybackslash}p{3.2cm}|ccccc@{}}
\toprule
& \multicolumn{3}{c}{\textbf{Stage 1: Generation Pre-Training}}
& & \\
\multirow{-2}{=}{\textbf{Training Stage}}
& \textbf{Phase I} & \textbf{Phase II} & \textbf{Phase III}
& \multirow{-2}{*}{\textbf{\makecell{Stage 2: Unified\\Mid-Training}}}
& \multirow{-2}{*}{\textbf{\makecell{Stage 3: Unified\\SFT}}} \\
\midrule

\multicolumn{6}{@{}l@{}}{\textbf{Optimization}} \\
Training steps
& 180K & 100K & 185K & 80K & 10.5K \\
Peak learning rate
& $2\times10^{-4}$ & $1\times10^{-4}$ & $1\times10^{-4}$
& $2\times10^{-5}$ & $2\times10^{-5}$ \\
Min learning rate
& $2\times10^{-4}$ & $1\times10^{-4}$ & $2\times10^{-5}$
& $2\times10^{-5}$ & $0$ \\
LR scheduler
& Constant & Constant & Cosine decay & Constant & Cosine decay \\
Sequence length
& 8,196 & 20,480 & 20,480 & 32,768 & 32,768 \\
Optimizer
& \multicolumn{5}{c}{AdamW ($\beta_1=0.9$, $\beta_2=0.95$, $\epsilon=10^{-8}$)} \\

\addlinespace[2pt]
\multicolumn{6}{@{}l@{}}{\textbf{Module settings}} \\
Understanding branch
& \frozen & \frozen & \frozen & \trainable & \trainable \\
Generation branch
& \trainable & \trainable & \trainable & \trainable & \trainable \\
\multicolumn{1}{@{}l|}{$\lambda_{\mathrm{AR}}:\lambda_{\mathrm{Flow}}:\lambda_{\mathrm{Perc}}$}
& $0:1:0$ & $0:1:0$ & $0.1:1:0.1$ & $0.1:1:0.1$ & $0.1:1:0.1$ \\
Generation resolution
& $256^2$--$1024^2$ & $512^2$--$4096^2$ & $512^2$--$4096^2$ & $512^2$--$4096^2$ & $512^2$--$4096^2$ \\

\addlinespace[2pt]
\multicolumn{6}{@{}l@{}}{\textbf{Data sampling ratio}} \\
Understanding data
& 0.00 & 0.00 & 0.00 & 0.30 & 0.30 \\
Text-to-image data
& 1.00 & 1.00 & 0.60 & 0.40 & 0.40 \\
Image-editing data
& 0.00 & 0.00 & 0.30 & 0.20 & 0.30 \\
Interleaved data
& 0.00 & 0.00 & 0.10 & 0.10 & 0.10 \\
\bottomrule
\end{tabular*}
\par
\endgroup

\caption{\textbf{Training recipe of SenseNova-U1.5.} \frozen{} denotes frozen modules; \trainable{} denotes trainable modules.}
\label{tab:u15-training-recipe}
\end{table*}

\subsection{Training Procedure}

SenseNova-U1.5 progressively builds native multimodal capabilities, beginning with generation pre-training, unified mid-training, and unified supervised fine-tuning (Stages 1–3), as summarized in Table~\ref{tab:u15-training-recipe}. These stages are followed by capability-specific learning (Stage 4) and multi-expert on-policy distillation (Stage 5).

\noindent\textbf{Stage 1: Generation Pre-Training.} Building on a pre-trained understanding branch, we randomly initialize the generation branch and train it via pixel-space flow matching, conditioned on representations from the frozen understanding branch. 
SenseNova-U1.5 increases the compute budget and introduces a dedicated native 4K training phase. Training begins with text-to-image data at resolutions ranging from \(256 \times 256\) to \(1024 \times 1024\). During this phase, we train for 180K steps with a constant learning rate of \(2 \times 10^{-4}\) and a sequence length of 8,196. We then extend to higher-resolution text-to-image data ranging from \(512 \times 512\) to \(4096 \times 4096\), thereby strengthening the model's native 4K generation capability. This phase runs for 100K steps with a constant learning rate of \(1 \times 10^{-4}\) and an increased sequence length of 20,480. In the final phase, we introduce image-editing and interleaved-generation tasks and train for an additional 185K steps, broadening the model's generative capabilities across diverse downstream scenarios. The training mixture comprises 60\% text-to-image, 30\% image-editing, and 10\% interleaved image-text data. We apply a cosine learning-rate schedule that decays from \(1 \times 10^{-4}\) to \(2 \times 10^{-5}\), while maintaining the sequence length at 20,480. Beginning in this phase, we augment the flow-matching objective with an LPIPS-based perceptual loss~\cite{zhang2018unreasonableeffectivenessdeepfeatures} weighted by 0.1 to improve the generation of fine-grained visual details and local visual coherence. Notably, we maintain this combined generation objective throughout the subsequent unified mid-training and supervised fine-tuning stages.

\noindent\textbf{Stage 2: Unified Mid-Training.}
We jointly optimize two branches, where shared attention facilitates cross-branch information exchange while preserving task-specific representations. To balance general multimodal competence with diverse generative capabilities, we construct a mixed corpus consisting of 30\% text-only and multimodal-understanding data, 40\% text-to-image data, 20\% image-editing data, and 10\% interleaved image-text data. This mixture exposes the model to complementary forms of perception, synthesis, editing, and multimodal interaction within a unified training stage.
We train the model for 80K steps with a maximum sequence length of 32,768 tokens and a constant learning rate of \(2 \times 10^{-5}\). The understanding and generation losses are weighted by 0.1 and 1.0, respectively. 
This weighting helps preserve the pretrained understanding capability while assigning greater optimization emphasis to the more challenging generative objectives, leading to stable joint training and more effective capability integration.

\noindent\textbf{Stage 3: Unified Supervised Fine-Tuning.}
We further fine-tune the model on a curated collection of high-quality instruction-following data, following a task composition similar to Stage 2. This stage further strengthens instruction adherence and consolidates the capabilities acquired during earlier training into a unified model. Training proceeds for 10.5K steps with a cosine learning-rate schedule, decaying from \(2 \times 10^{-5}\) to 0, while retaining the same loss coefficients as in Stage 2 to preserve the balance between understanding and generation objectives.

\begin{figure*}[t]
    \centering
    \includegraphics[width=0.99\textwidth]{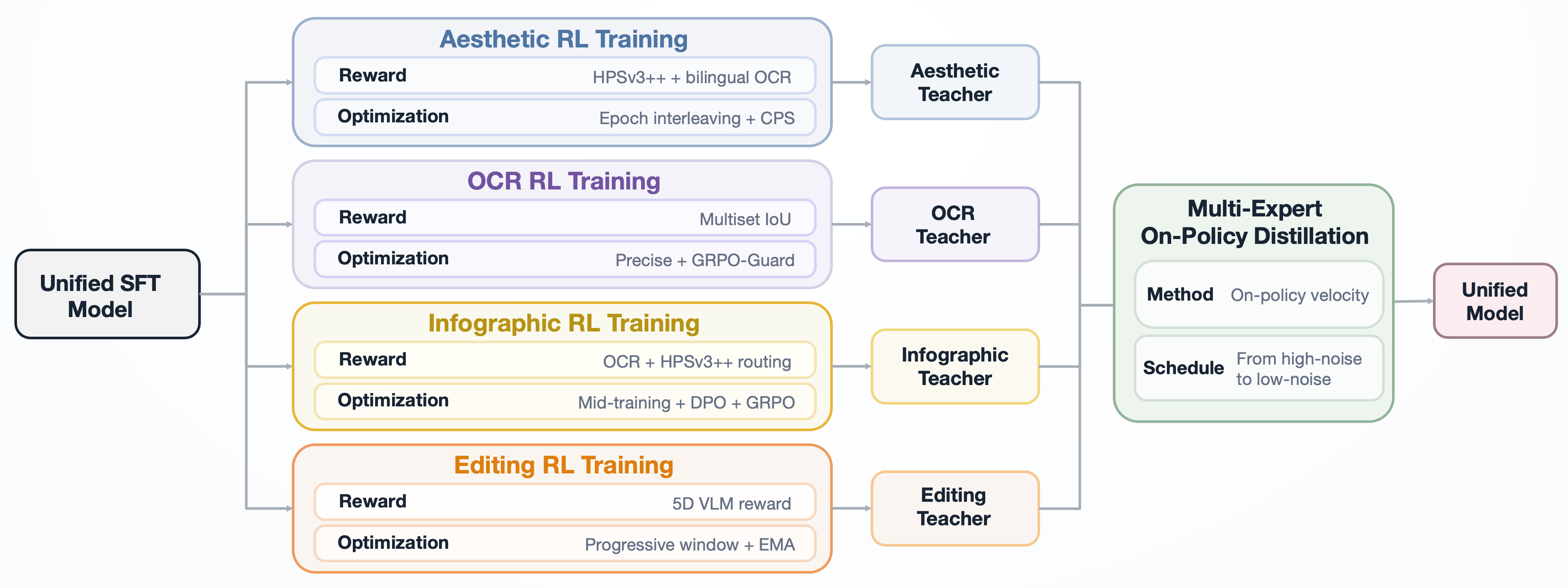}
    \caption{\textbf{Post-training process of SenseNova-U1.5}, including multi-expert reinforcement learning and on-policy distillation.}
    \label{fig:rl-pipeline}
\end{figure*}

\noindent\textbf{Stage 4: Multi-Expert Reinforcement Learning.}
In Figure~\ref{fig:rl-pipeline}, we train four experts for aesthetics, text rendering, infographic generation, and image editing, each with task-specific data, rewards, sampling, and regularization.

\noindent\textbf{(1) Aesthetic Expert.}
Optimizing visual preference alone can improve global appearance at the expense of text legibility. We therefore train the aesthetic expert by interleaving aesthetic-preference and typography data across epochs. Samples from each partition are routed to task-specific rewards: HPSv3++~\cite{liu2026hpsv3pp} evaluates perceptual quality and prompt-image alignment, while a bilingual OCR reward based on PaddleOCR~\cite{du2020ppocr} measures text fidelity. This routing avoids mixing rewards with different scales and semantics, while preserving typographic accuracy during preference optimization.

For each prompt, we generate 16 candidates using a 30-step trajectory with guidance scale 4.0 and timestep shift 3. We adopt coefficients-preserving sampling (CPS) with $\eta=0.7$~\cite{wang2025cps} to introduce stochastic exploration while reducing finite-step artifacts from conventional SDE sampling. At each update, we uniformly sample a contiguous five-step window from the first ten trajectory steps, focusing optimization on early denoising stages that largely determine global content and composition. Training uses dynamic resolutions and aspect ratios, a constant learning rate of $2\times10^{-5}$, and a reference-policy KL coefficient of $0.01$. To reduce this reward-induced drift, we freeze the final generation Transformer block, the flow-matching output head, and its output-normalization layer during the RL process.

\noindent\textbf{(2) OCR Expert.}
The OCR expert specializes the model for accurate bilingual text rendering. For each prompt, we extract the intended in-image text as the target and apply PaddleOCR~\cite{du2020ppocr} to recognize the generated content. We then compute a normalized multiset intersection-over-union score between the recognized and target text, matching English at the word level and Chinese at the character level. Compared with exact matching, this graded reward better captures partially correct generations while remaining sensitive to omissions, duplications, and rendering errors.

We generate 16 candidates per prompt using a 30-step trajectory with guidance scale 4.0 and timestep shift 3, and sample a contiguous five-step optimization window from the first ten steps. Rollouts use Precise sampling with $\eta=1.5$~\cite{zou2026precise}, while GRPO-Guard~\cite{wang2025grpoguard} stabilizes optimization under sparse high OCR rewards. Training uses dynamic resolutions and aspect ratios, a constant learning rate of $2\times10^{-5}$, and a reference-policy KL coefficient of $0.02$. We also freeze the final generation Transformer block, flow-matching output head, and output-normalization layer during RL.

\noindent\textbf{(3) Editing Expert.}
Effective image editing requires balancing three objectives: instruction adherence, preservation of unedited content, and the visual quality of edited regions~\cite{wu2026editrewardhumanalignedrewardmodel,wang2026rationalrewardsreasoningrewardsscale}. Under-editing can leave instructions only partially fulfilled, while over-editing may alter content that should remain unchanged. The editing expert must therefore learn what to modify, what to preserve, and how to integrate the requested changes naturally into the source image.

The editing expert is trained with a constant learning rate of learning rate of $4\times10^{-5}$, KL coefficient of $0.01$, rollout size of 24, and noise scale of $0.7$, with an exponential moving average (EMA). We avoid SDE-based trajectory sampling, as excessive stochastic perturbations can introduce residual noise and visual artifacts. Under this setup, we further introduce two editing-specific strategies for targeted reward supervision and more effective optimization.

\textit{(i) Multi-dimensional reward.}
We utilize a VLM-based reward framework that evaluates editing quality along five dimensions: instruction fulfillment, edit execution, overall visual quality, text-editing quality when applicable, and preservation of unedited regions. These complementary signals provide targeted supervision for distinct failure modes while jointly balancing edit accuracy, visual fidelity, and content preservation.

\textit{(ii) Progressive training.}
We adopt a progressive sliding-window strategy, shifting optimization from early stages that establish semantics to later stages that refine textures. This coarse-to-fine progression first secures the intended structural change and then improves local fidelity and visual integration with the surrounding content.

\noindent\textbf{Infographic Expert.}
Infographic generation requires accurate text rendering, coherent dense layouts, clear information hierarchies, and strong overall visual quality. The infographic expert first undergoes a complete training process from infographic-oriented mid-training to task-specific RL, aiming to achieve more substantial improvements in small-text rendering and related capabilities. During mid-training, we introduce high-quality real and synthetic data and rebalance both the infographic subcategories and the mixture of infographic and other categories of data. This further improves small-text rendering, complex layout handling, information organization, and overall visual design quality.

Here, we further refine the infographic expert in three stages. The first improves text rendering using the same training data and text-fidelity reward as the OCR expert. The second applies direct preference optimization (DPO)~\cite{li2026lineardpolineardirectpreference} to curated preference pairs to enhance overall visual quality, using a constant learning rate of \(5\times10^{-6}\), a DPO loss coefficient of \(\beta=10\), and one timestep sampled from steps 1--20 at each update. The final stage alternates text-rendering and aesthetic data, routing them separately to the text-fidelity reward and HPSv3++~\cite{liu2026hpsv3pp}, rather than merging the two scores.

The first and final stages use group relative policy optimization (GRPO), generating 16 candidates per prompt over a 30-step trajectory with coefficients-preserving sampling (CPS)~\cite{wang2025cps}. Classifier-free guidance is used only during rollout collection for reward evaluation, while the policy objective is computed from conditional predictions alone. Both stages use a constant learning rate of \(2\times10^{-5}\) and a reference-policy KL coefficient of \(0.01\).

\noindent\textbf{Stage 5: Multi-Expert On-Policy Distillation.}
In Figure~\ref{fig:rl-pipeline}, we apply on-policy velocity-field distillation to consolidate four specialized experts for aesthetic quality, text rendering, infographic generation, and image editing into a single unified model. Each training sample is hard-routed to the frozen expert corresponding to its capability. Given a text-only or text-image condition \(\mathbf{c}\) from capability \(m\), the student first generates its own trajectory \(\{\hat{\mathbf{x}}_{\theta,t}\}_{t=0}^1\). The student and routed expert are then evaluated at the same student-generated state, timestep, and condition:

\[
\mathcal{L}_{\mathrm{OPD}}
=\mathbb{E}\!\left[
\left\|
\mathbf{v}_{\theta}\!\left(\operatorname{sg}(\hat{\mathbf{x}}_{\theta,t}),t,\mathbf{c}\right)
-\mathbf{v}_m\!\left(\operatorname{sg}(\hat{\mathbf{x}}_{\theta,t}),t,\mathbf{c}\right)
\right\|_2^2
\right].
\]

Here, \(\operatorname{sg}\) denotes stop-gradient, preventing gradients from propagating through the full sampling trajectory. Each epoch uses a single capability partition and its paired frozen expert, rotating in a fixed order.

Both text-to-image generation and image editing use fixed 30-step deterministic ODE denoising steps, with one queried timestep per trajectory. To learn both high-noise states that establish global structure and low-noise states that determine local details, we progressively shift the query distribution over training. At the \(n\)-th epoch, the continuous query location and its discrete 0-based index are determined by
$
t_e\sim\operatorname{Beta}\!\left(2+3n/N,\,5-3n/N\right)
$,
where \(N\) is the total number of epochs. The distribution therefore moves smoothly from \(\operatorname{Beta}(2,5)\), which emphasizes the early high-noise portion of the trajectory, to \(\operatorname{Beta}(5,2)\), which emphasizes late low-noise states and hence fine details and text fidelity.
To align training and inference, we directly optimize the velocity under classifier-free guidance (CFG). We use scale $s=4$ for the aesthetic, OCR, and infographic experts, with global-norm clipping, and scale $s=1$ for the editing expert.

For text-to-image generation, we first choose a target area from \(
\mathcal A=\left\{1024^2,1536^2,2048^2,3072^2,4096^2\right\}
\), and a target aspect ratio from \(\left\{1{:}1,\ 16{:}9,\ 9{:}16,\ 4{:}3,\ 3{:}4,\ 3{:}2,\ 2{:}3,\ 1{:}2,\ 2{:}1\right\}\).
The final width and height are computed by preserving the area and rounded to the nearest multiple of 32. For image editing, the generation resolution follows that of the source image, with only rounding to the nearest multiple of 32 pixels.

The student is trained for 800 optimizer steps with a global batch size of 128 and gradient accumulation of 1, using 25,600 samples per domain. We use AdamW with learning rate \(2\times10^{-5}\), \(\beta_1=0.9\), \(\beta_2=0.999\), weight decay \(10^{-4}\), maximum gradient norm 1, and BF16 mixed precision. The understanding branch, the final three generation layers, and the generation output head are frozen. The entire training process adopts a 30-step trajectory, but terminates immediately after the sampled query transition to avoid computing the unused trajectory suffix.

\subsection{Reward Modeling}

\textbf{Aesthetic Reward.}
Aesthetic quality is difficult to capture with hand-crafted objectives, as it jointly reflects composition, visual fidelity, semantic consistency, stylistic coherence, and broader human preference. We therefore adopt HPSv3++~\cite{liu2026hpsv3pp} as the preference reward $R_\mathrm{HPSv3++}$. Its capability- and iteration-aware training makes it particularly suitable for RL post-training, where the model distribution continuously shifts as generation quality improves. This provides more robust and adaptive preference supervision than scorers trained on a fixed generation distribution, helping the model pursue perceptual improvements without overfitting to a narrow reward landscape.

\textbf{OCR Reward.}
Preference rewards alone are insufficient for typography, where visually plausible images may still contain missing, duplicated, or malformed text. We therefore introduce an explicit OCR reward. Target transcriptions are extracted from text spans specified in the prompt, while generated images are recognized using PaddleOCR~\cite{du2020ppocr}. After normalizing case, punctuation, and whitespace, English text is tokenized into words and Chinese text into characters. Let \(g_t\) and \(o_t\) denote the multiplicities of token \(t\) in the target and OCR transcription, respectively. We define:
\begin{equation}
    R_{\mathrm{ocr}}
    =
    \frac{
        \sum_t \min(g_t,o_t)
    }{
        \sum_t \max(g_t,o_t)
    }.
\end{equation}
This multiset IoU rewards correct token occurrences while symmetrically penalizing omissions, hallucinations, and errors in repeated text. It is also robust to unreliable OCR reading order in dense or multi-region layouts. During post-training, aesthetic and OCR prompts use their respective rewards and are interleaved at the epoch level, avoiding the need to combine heterogeneous reward scales within each sample.

\textbf{Infographic Reward.}
Note that we do not introduce an additional infographic-specific reward model. Instead, the infographic expert reuses the OCR and aesthetic rewards above through stage-conditioned and task-conditioned routing. Here, the first training stage is optimized exclusively with the OCR reward for warmup:
\begin{equation}
    R_{\mathrm{info}}^{warmup}(\hat{\mathbf{x}}_{\theta},\mathbf{c})
    =
    R_{\mathrm{ocr}}(\hat{\mathbf{x}}_{\theta},\mathbf{c}),
\end{equation}
where \(\hat{\mathbf{x}}_{\theta}\) is a generated image and \(\mathbf{c}\) is its corresponding text prompt.
In the final training stage, let \(m\in\{\mathrm{ocr},\mathrm{aesthetic}\}\) identify the prompt partition. The reward is assigned as follows:
\begin{equation}
    R_{\mathrm{info}}^{final}(\hat{\mathbf{x}}_{\theta},\mathbf{c},m)
    =
    \begin{cases}
        R_{\mathrm{ocr}}(\hat{\mathbf{x}}_{\theta},\mathbf{c}),
        & m=\mathrm{ocr},\\
        R_{\mathrm{HPSv3++}}(\hat{\mathbf{x}}_{\theta},\mathbf{c}),
        & m=\mathrm{aesthetic}.
    \end{cases}
\end{equation}
The OCR and HPSv3++ scores are therefore not summed or normalized against one another. This routing preserves their distinct scales and semantics while providing complementary supervision for textual fidelity and perceptual quality. 

\textbf{Editing Reward.} 
Reinforcement learning for image editing requires reward signals that assess not only whether the requested change is successfully completed, but also whether the result is visually convincing and unrelated source content is preserved. A single holistic score can obscure these distinct failure modes, allowing strength in one aspect to compensate for critical errors in another. We therefore employ a set of dimension-specific rewards. \textit{(i) Instruction fulfillment} measures the semantic correctness of the edited result and the completeness of the requested changes. \textit{(ii) Edit execution quality} evaluates the visual quality of the target region and how naturally it integrates with the surrounding context. \textit{(iii) Overall visual quality} captures global technical quality and coherence with the intended style. When visible text modification is required, \textit{(iv) text-editing quality} further assesses textual correctness and rendering quality. \textit{(v) Unedited-region preservation} compares the source and edited images and penalizes unintended semantic, structural, or visual changes outside the target region. For criteria with multiple sub-dimensions, we retain the minimum sub-dimension score to expose the weakest aspect. After normalization, the applicable scores are aggregated as
\begin{equation}
    R_{\mathrm{edit}} = \min\!\left(
        \left\{R_{\mathrm{inst}}, R_{\mathrm{exec}},
        R_{\mathrm{visual}}, R_{\mathrm{pres}}\right\}
        \cup \left\{R_{\mathrm{text}}\ \middle|\
        \mathrm{text\ edit\ is\ applicable}\right\}
    \right).
\end{equation}
The text-editing criterion is omitted for instructions that do not involve text modification, while the edit-execution score is set to zero when no requested change is visibly realized. This bottleneck-style aggregation emphasizes the weakest aspect of each edit and prevents strong performance in one dimension from masking critical failures in another. As a result, the reward remains sensitive to incomplete edits, unintended modifications, and local quality defects, providing more balanced and reliable supervision for image-editing RL during the training process.

\section{Data Construction}

\begin{figure}[t]
  \centering
  \includegraphics[width=0.99\textwidth]{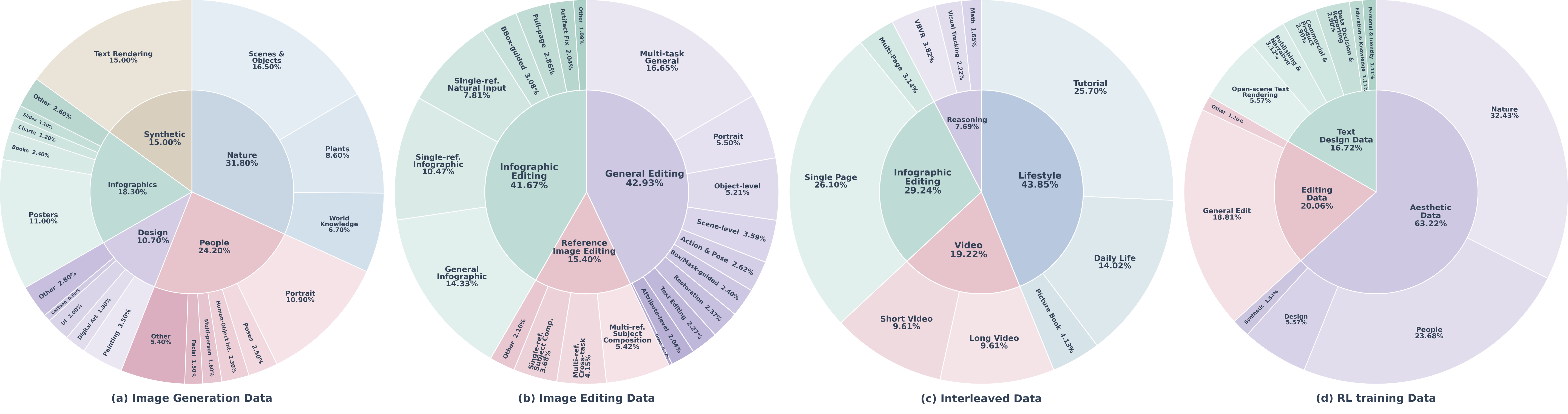}
  \caption{\textbf{Training corpus of SenseNova-U1.5.} From left to right, the charts show the hierarchical composition of the \textit{Image Generation}, \textit{Image Editing}, \textit{Interleaved}, and \textit{RL Training} datasets. The inner ring denotes major data categories and their proportions, while the outer ring further decomposes them into fine-grained subclasses. Together, these distributions highlight the broad coverage of natural and synthetic imagery, diverse editing scenarios, interleaved image-text and video content, and post-training data.}
  \label{fig:data_dist}
\end{figure}

In Figure~\ref{fig:data_dist}, SenseNova-U1.5 substantially expands the training corpus of SenseNova-U1~\cite{diao2026sensenovau1}, with an emphasis on data diversity, visual quality, high-resolution generation, complex instruction following, and fine-grained image editing. Beyond increasing data scale, we systematically improve data curation through quality-aware filtering, distribution rebalancing, stronger image-text alignment, and targeted synthesis for underrepresented capabilities. 

\subsection{Image Generation Data}
The image-generation corpus combines large-scale real-world image--text pairs, carefully curated public and proprietary sources, and high-quality synthetic data. Compared with SenseNova-U1, we introduce approximately 59M additional text--image pairs collected from 78 sources, substantially broadening the training distribution across general scenes, human-centric content, objects and materials, text-rich imagery, infographics, and other specialized visual domains. Importantly, high-resolution samples constitute a substantial fraction of the overall training mixture, with approximately 88.2\% of the effective training volume exceeding $1024^2$ resolution and 64.4\% exceeding $2048^2$.
This exposes the model to rich visual statistics at multiple spatial scales, providing supervision not only for globally coherent scene composition and layout, but also for fine-grained local structures, textures, materials, and other high-frequency visual details.

To improve instruction alignment, captions are constructed at multiple levels of granularity, ranging from long detailed descriptions to concise captions and lightweight semantic tags. We further strengthen bilingual Chinese-English coverage, particularly for text-rich images, and explicitly verify the consistency between the language and text specified in the prompt and that rendered in the image. To complement real-world data, we collect web-scale data and also introduce targeted synthesis for rare concepts, complex layouts, and text-intensive designs. Both synthetic and real samples are processed through a shared filtering pipeline that jointly evaluates perceptual quality, prompt fidelity, text correctness, layout quality, and visual diversity, ensuring a high-quality and well-balanced training distribution.

\subsection{Image Editing Data}
Our image-editing corpus contains approximately 38M examples and is designed to cover four complementary settings: general image editing, infographic editing, reference-conditioned editing, and spatially controlled editing. Together, these settings span a broad range of editing scenarios, including localized attribute modification, object insertion and removal, scene-level transformation, pose and action editing, restoration and enhancement, text and layout manipulation, reference-guided customization, and region-controlled operations. To improve robustness beyond fixed instruction patterns, we diversify both the underlying visual content and the formulation, granularity, and complexity of editing instructions. For challenging multi-target or multi-constraint edits, we further incorporate structured prompt-enhancement and chain-of-thought examples that explicitly characterize the editing intent, target regions, desired attributes, spatial or semantic constraints, and content that should remain unchanged. This structured supervision reduces instruction ambiguity and provides stronger guidance for precise, faithful, and preservation-aware editing.

The corpus comprises $\sim43\%$ general editing, $\sim42\%$ spatially controlled infographic editing examples, together with a substantial $\sim15\%$ reference-conditioned subset that supports both single- and multi-reference editing. Infographic data combines programmatically rendered source--target pairs with model-generated examples, enabling precise manipulation of text, visual elements, layout structures, and overall styles. Reference-conditioned data emphasizes subject and identity preservation, attribute transfer, compositional integration, and controlled transformations across viewpoint, pose, geometry, and illumination, with each example containing up to ten reference images. Spatially controlled data further introduces explicit region-level guidance through bounding boxes and visual markers, supporting more precise and interactive manipulation. Across all editing settings, data curation jointly evaluates instruction fulfillment, target quality, reference or identity fidelity when applicable, and preservation of content outside the intended edit region.

\subsection{Interleaved Data}
The majority of the corpus is drawn from lifestyle-oriented trajectories ($\sim44\%$), including procedural tutorials, everyday scenarios, and picture-book narratives. These examples provide rich supervision for maintaining long-range semantic coherence, visual consistency, and cross-modal dependencies across alternating text and image segments. Infographic data contributes another $\sim29\%$, emphasizing structured composition, dense textual content, and coordinated generation of linguistic and visual elements. Video-derived sequences account for approximately $\sim19\%$, introducing temporal evolution, state transitions, and cross-frame dependencies. The remaining $\sim8\%$ consists of reasoning-intensive examples augmented with explicit intermediate reasoning for complex multi-step understanding and generation.

Rather than treating these sources as isolated task collections, we convert them into a unified trajectory format in which textual and visual states are progressively interleaved~\cite{cui2025emu35nativemultimodalmodels,xing2026wan}. This unified formulation exposes the model to a broad spectrum of multimodal dependencies, ranging from local image--text grounding to long-range semantic progression, temporal continuity, and reasoning-conditioned generation across multiple steps. It is worth noting that all trajectories are constructed through a shared pipeline that combines source preprocessing, domain-specific transformation or synthesis, and trajectory-level validation. The resulting corpus is jointly filtered for linguistic validity, visual quality, cross-modal consistency, and the coherence and correctness of the complete multimodal progression.

\subsection{RL Training Data}

\textbf{Aesthetic, OCR, and Infographic Samples.}
For aesthetic samples, we construct approximately 280K prompts by combining HPSv3++~\cite{liu2026hpsv3pp}, Pick-a-Pic~\cite{kirstain2023pickapic}, bilingual prompts expanded with Cosmos3PE~\cite{nvidia2026cosmos3}, and internally generated prompts covering diverse subjects, styles, compositions, and spatial configurations. Multiple images are generated for each prompt and scored with HPSv3++; prompt groups with weak reward variation are removed to retain examples that provide more discriminative relative supervision. The OCR corpus contains approximately 60K bilingual text-rendering prompts with balanced Chinese and English coverage. It includes approximately 20K short-text prompts collected from Flow-GRPO~\cite{liu2025flowgrpo}, which are further rewritten and expanded into approximately 40K longer prompts with denser text and more complex rendering requirements; the specified text is extracted as the reference transcription for online OCR reward computation. For infographic samples, the first GRPO stage reuses the OCR corpus, while the later GRPO stage jointly iterates OCR and aesthetic prompts according to their task-specific rewards. We further introduce approximately 120K filtered preference pairs derived from the Linear-DPO corpus~\cite{li2026lineardpolineardirectpreference} for DPO, with additional balancing over portrait/non-portrait content and Chinese/English prompts to improve overall visual-preference alignment.

\noindent\textbf{Image Editing Samples.}
The editing expert is trained on approximately 120K examples, with local and global editing accounting for roughly 80\% and 20\% of the corpus, respectively. Local editing covers infographic and document manipulation, scene- and object-level edits, human action and pose adjustment, portrait retouching, text replacement, image restoration and enhancement, and bounding-box- or mask-conditioned editing, while global editing focuses on coherent image-level transformations. Chinese, English, and mixed-language instructions are included, with instruction lengths balanced across predefined ranges and training resolutions spanning from \(512\times512\) to \(2048\times2048\). Data curation follows three stages. We first remove corrupted, duplicated, low-resolution, or visually degraded samples and retain a diverse set of common aspect ratios. We then verify that the entities, text, spatial regions, and visual attributes specified by each instruction are grounded in the source image and that the target image faithfully realizes the requested modification. Finally, both source and edited images are evaluated for aesthetics, technical quality, and structural and textural richness, with conservative pairwise aggregation exposing deficiencies in either image. The retained samples are then rebalanced across task categories and resolution levels to form a diverse and challenging distribution.

\section{Experiments}

\subsection{General Understanding}

\begin{table}[t]
\centering
\scriptsize
\caption{\textbf{Quantitative evaluation results on multimodal and language understanding benchmarks}.
Note that $^{*}$ denotes results reproduced using VLMEvalKit~\cite{duan2024vlmevalkit}. SenseNova-U1, SenseNova-U1.5, and Qwen3-VL are all initialized from Qwen3 LLM weights~\cite{yang2025qwen3}. Besides, Gemma4-12B~\cite{gemma42026} represents a recent state-of-the-art encoder-free model for visual understanding.}
\label{tab:multimodal_text_understanding_results}
\vspace{1em}
\setlength{\tabcolsep}{1pt}
\begin{tabularx}{\textwidth}{@{}l*{2}{>{\centering\arraybackslash}X}|*{3}{>{\centering\arraybackslash}X}@{}}
\toprule
\textbf{Benchmark}
& \makecell{\textbf{SenseNova-U1.5}\\\textbf{8B-Think}}
& \makecell{\textbf{SenseNova-U1}\\\textbf{8B-Think}}
& \makecell{\textbf{Qwen3VL}\\\textbf{8B-Think}}
& \makecell{\textbf{Qwen3.5}\\\textbf{9B}}
& \makecell{\textbf{Gemma4}\\\textbf{12B}} \\
\midrule

\multicolumn{6}{c}{\textit{STEM \& Reasoning}} \\
\midrule
MMMU~\cite{yue2024mmmu} 
& 73.86 & 74.78 & 74.10 & 78.40 & 64.67\rlap{$^{*}$} \\

MMMU-Pro~\cite{yue2025mmmu} 
& 66.47 & 67.69 & 60.40 & 70.10 & 69.10 \\

MathVista$_{\text{mini}}$~\cite{lu2023mathvista} 
& 85.85 & 84.20 & 81.40 & 85.70 & 74.70\rlap{$^{*}$} \\

MathVision~\cite{wang2024measuring} 
& 73.55 & 75.82 & 62.70 & 78.90 & 79.70 \\

\midrule
\multicolumn{6}{c}{\textit{General VQA}} \\
\midrule
MMBench-EN~\cite{liu2024mmbench} 
& 90.47 & 90.25 & 87.50 & 90.10 & 83.85\rlap{$^{*}$} \\

MMStar~\cite{chen2024we} 
& 77.53 & 78.27 & 75.30 & 79.70 & 70.13\rlap{$^{*}$} \\

\midrule
\multicolumn{6}{c}{\textit{OCR}} \\
\midrule
InfoVQA~\cite{mathew2022infographicvqa} 
& 82.51 & 82.46 & 86.00 & 90.76 & 88.40 \\

OCRBench-v2~\cite{fu2024ocrbench} 
& 59.07 & 61.30 & 61.55 & 66.54 & 67.46\rlap{$^{*}$} \\

AI2D~\cite{hiippala2021ai2d} 
& 92.03 & 91.74 & 84.90 & 90.20 & 84.00\rlap{$^{*}$} \\

OCRBench~\cite{liu2024ocrbench} 
& 82.46 & 82.10 & 81.90 & 89.20 & 76.90\rlap{$^{*}$} \\

\midrule
\multicolumn{6}{c}{\textit{Hallucination}} \\
\midrule
HallusionBench~\cite{guan2024hallusionbench} 
& 68.53 & 67.75 & 65.40 & 69.30 & 66.43\rlap{$^{*}$} \\

\midrule
\multicolumn{6}{c}{\textit{Visual Reasoning}} \\
\midrule
BabyVision~\cite{chen2026babyvision} 
& 25.52 & 25.00 & 17.78 & 25.80 & 22.43\rlap{$^{*}$} \\

TiR~\cite{li2025tir} 
& 28.68 & 28.15 & 22.30 & 31.90 & 27.32\rlap{$^{*}$} \\

\midrule
\multicolumn{6}{c}{\textit{Knowledge}} \\
\midrule
MMLU-Pro~\cite{wang2024mmlu} & 86.67 & 81.44 & 77.30 & 82.50 & 77.20 \\
MMLU-Redux~\cite{gema2025we} & 86.33 & 87.61 & 88.80 & 91.10 & 88.33\rlap{$^{*}$} \\
C-Eval~\cite{huang2023c} & 90.41 & 84.40 & 83.88 & 88.20 & 84.23\rlap{$^{*}$} \\
SuperGPQA~\cite{du2025supergpqa} & 49.55 & 49.67 & 51.20 & 58.20 & 48.43\rlap{$^{*}$} \\
\midrule
\multicolumn{6}{c}{\textit{Instruction Following}} \\
\midrule
IFEval~\cite{zhou2023instruction} & 93.35 & 91.13 & 83.20 & 91.50 & 97.20 \\
IFBench~\cite{zhang2025if} & 69.00 & 67.01 & 29.93 & 64.50 & 74.00 \\

\bottomrule
\end{tabularx}
\end{table}
\textbf{Multimodal Understanding.}
In Table~\ref{tab:multimodal_text_understanding_results}, SenseNova-U1.5 shows consistently strong performance across diverse multimodal benchmarks, covering STEM reasoning, general VQA, OCR, hallucination, and visual reasoning. Compared with SenseNova-U1~\cite{diao2026sensenovau1}, U1.5 preserves or improves performance on most benchmarks while substantially extending its image-generation and editing capabilities, indicating that stronger generative modeling does not come at the expense of visual understanding. In particular, U1.5 achieves competitive results on MMMU, MathVista, MMBench, MMStar, AI2D, OCRBench, HallusionBench, BabyVision, and TiR, and remains comparable to or better than strong modular baselines such as Qwen3-VL~\cite{Qwen3-VL} on many tasks. Notably, despite adopting an encoder-free architecture, U1.5 also compares favorably with the recent encoder-free Gemma4-12B~\cite{gemma42026} across various visual understanding evaluations, highlighting the effectiveness of native multimodal unification for general-purpose perception and reasoning.

\textbf{Language Understanding.}
Beyond multimodal capability, SenseNova-U1.5 retains strong language understanding and instruction-following performance after unified multimodal training. It achieves particularly strong results on MMLU-Pro and C-Eval, reaching 86.67 and 90.41, respectively, while remaining competitive on MMLU-Redux and SuperGPQA. More importantly, U1.5 substantially improves instruction-following performance over SenseNova-U1, achieving 93.35 on IFEval and 69.00 on IFBench. This suggests that visual generation and editing capabilities introduced in U1.5 do not significantly dilute its underlying language competence. Instead, the model maintains a strong text reasoning foundation while supporting a substantially broader multimodal capability space within a single unified model.

\begin{table}[t]
\centering
\caption{\textbf{Quantitative evaluation results on Qwen-Image-Bench-EN}.
Results are reported on the English subset.
\textit{\# Params} denotes the number of parameters in the generation component;
\textit{A} in this column denotes activated parameters during inference.}
\label{tab:qwen_image_bench_en_results}

\renewcommand{\arraystretch}{0.95}
\setlength{\tabcolsep}{3pt}
\small

\begin{adjustbox}{max width=0.95\textwidth}
\begin{tabular}{lc|ccccc|c}
\toprule
\textbf{Model}
& \textbf{\# Params}
& \textbf{Quality}
& \textbf{Aesthetics}
& \textbf{Alignment}
& \shortstack{\textbf{Real-world}\\\textbf{Fidelity}}
& \shortstack{\textbf{Creative}\\\textbf{Generation}}
& \textbf{Overall}$\uparrow$ \\
\midrule

\multicolumn{8}{c}{\textit{Closed-source Models}} \\
\midrule
GPT-Image-2~\cite{gpt_image_2}
& --
& 59.09
& 68.48
& 65.78
& 59.40
& 75.34
& 65.23 \\

GPT-Image-1.5~\cite{GPT-Image-1.5}
& --
& 55.78
& 62.87
& 61.39
& 55.86
& 67.06
& 60.42 \\

Nano-Banana-2.0~\cite{nanobanana2}
& --
& 54.86
& 62.63
& 61.11
& 54.66
& 64.49
& 59.59 \\

Nano-Banana-Pro~\cite{deepmind_gemini3proimage_2025}
& --
& 55.30
& 61.38
& 60.30
& 55.91
& 64.54
& 59.33 \\

Seedream 5.0~\cite{Seedream5}
& --
& 54.01
& 59.96
& 58.63
& 53.86
& 63.64
& 57.80 \\

Seedream 4.5~\cite{seedream45}
& --
& 54.05
& 60.11
& 57.44
& 51.55
& 59.82
& 56.95 \\

Qwen-Image2-Pro~\cite{zhao2026qwenimage20technicalreport}
& --
& 55.16
& 60.36
& 57.86
& 53.06
& 63.59
& 57.90 \\

\midrule
\multicolumn{8}{c}{\textit{Open-source Models}} \\
\midrule
\textbf{SenseNova-U1.5 (w/ PE)}
& \textbf{8B}
& 54.74
& 63.72
& 61.79
& 53.40
& 67.89
& \textbf{60.22} \\

LLaDA-Image~\cite{chen2026llada}
& 6B
& 53.22
& 58.22
& 54.77
& 43.90
& 51.09
& 53.53 \\

\textbf{SenseNova-U1.5}
& \textbf{8B}
& 51.27
& 55.19
& 54.64
& 48.67
& 51.87
& \textbf{52.82} \\

Qwen-Image-2512~\cite{wu2025qwenimagetechnicalreport}
& 20B
& 51.84
& 54.40
& 51.44
& 47.80
& 47.75
& 51.32 \\

Boogu-Image 0.1 Base~\cite{chen2026boogu}
& 10B
& 50.38
& 53.90
& 52.62
& 46.54
& 47.52
& 51.00 \\


SenseNova-U1~\cite{diao2026sensenovau1}
& 8B
& 47.30
& 50.30
& 50.58
& 43.13
& 46.15
& 48.28 \\


\bottomrule
\end{tabular}
\end{adjustbox}
\end{table}

\begin{table}[!t]
\centering
\caption{\textbf{Quantitative evaluation results on Qwen-Image-Bench-ZH}.
Results are reported on the Chinese subset.
\textit{\# Params} denotes the number of parameters in the generation component;
\textit{A} in this column denotes activated parameters during inference.}
\label{tab:qwen_image_bench_zh_results}

\renewcommand{\arraystretch}{0.95}
\setlength{\tabcolsep}{3pt}
\small

\begin{adjustbox}{max width=0.95\textwidth}
\begin{tabular}{lc|ccccc|c}
\toprule
\textbf{Model}
& \textbf{\# Params}
& \textbf{Quality}
& \textbf{Aesthetics}
& \textbf{Alignment}
& \shortstack{\textbf{Real-world}\\\textbf{Fidelity}}
& \shortstack{\textbf{Creative}\\\textbf{Generation}}
& \textbf{Overall}$\uparrow$ \\
\midrule

\multicolumn{8}{c}{\textit{Closed-source Models}} \\
\midrule
GPT-Image-2~\cite{gpt_image_2}
& --
& 58.65
& 67.53
& 65.85
& 57.38
& 75.23
& 64.69 \\

Nano-Banana-2.0~\cite{nanobanana2}
& --
& 54.77
& 61.08
& 62.40
& 54.28
& 67.05
& 59.82 \\

GPT-Image-1.5~\cite{GPT-Image-1.5}
& --
& 55.14
& 60.88
& 61.72
& 53.95
& 66.35
& 59.65 \\

Nano-Banana-Pro~\cite{deepmind_gemini3proimage_2025}
& --
& 55.67
& 60.26
& 61.25
& 54.07
& 66.23
& 59.45 \\

Seedream 5.0~\cite{Seedream5}
& --
& 52.55
& 58.40
& 58.90
& 51.92
& 65.29
& 57.22 \\

Seedream 4.5~\cite{seedream45}
& --
& 54.41
& 58.72
& 57.31
& 51.69
& 60.64
& 56.78 \\

Qwen-Image2-Pro~\cite{zhao2026qwenimage20technicalreport}
& --
& 54.39
& 58.67
& 59.28
& 51.83
& 64.94
& 57.84 \\

\midrule
\multicolumn{8}{c}{\textit{Open-source Models}} \\
\midrule
\textbf{SenseNova-U1.5 (w/ PE)}
& \textbf{8B}
& 54.42
& 62.21
& 62.61
& 52.39
& 68.56
& \textbf{60.13} \\

LLaDA-Image~\cite{chen2026llada}
& 6B
& 52.92
& 56.92
& 54.87
& 44.86
& 52.10
& 53.38 \\

\textbf{SenseNova-U1.5}
& \textbf{8B}
& 51.37
& 54.33
& 54.66
& 47.13
& 51.31
& \textbf{52.43} \\

Qwen-Image-2512~\cite{wu2025qwenimagetechnicalreport}
& 20B
& 51.76
& 54.74
& 52.72
& 47.00
& 50.19
& 52.06 \\

Boogu-Image 0.1 Base~\cite{chen2026boogu}
& 10B
& 50.41
& 53.14
& 52.91
& 45.42
& 48.62
& 50.96 \\

Qwen-Image~\cite{wu2025qwenimagetechnicalreport}
& 20B
& 48.44
& 52.25
& 50.72
& 43.16
& 47.30
& 49.23 \\


SenseNova-U1~\cite{diao2026sensenovau1}
& 8B
& 45.79
& 47.07
& 47.78
& 42.53
& 43.88
& 45.99 \\

\bottomrule
\end{tabular}
\end{adjustbox}
\end{table}

\subsection{Image Generation}

\textbf{General Generation.}
We evaluate general text-to-image generation on Qwen-Image-Bench~\cite{li2026qwen}, GenEval~\cite{ghosh2023geneval}, GenEval2~\cite{kamath2025geneval2}, DPG-Bench~\cite{hu2024ella}, and OneIG-Bench~\cite{chang2025oneig}, covering visual quality, prompt alignment, compositional generation, dense instruction following, text rendering, style, and diversity. 


\paragraph{Qwen-Image-Bench.}
In Tables~\ref{tab:qwen_image_bench_en_results} and~\ref{tab:qwen_image_bench_zh_results}, SenseNova-U1.5 delivers strong overall performance on both the English and Chinese subsets of Open-Image-Bench. With prompt enhancement, it achieves scores of 60.22 and 60.13, respectively, reaching the best overall performance among the evaluated open-source models. Even without prompt enhancement, SenseNova-U1.5 consistently improves over SenseNova-U1 and remains competitive with larger open-source baselines, demonstrating strong bilingual generation quality and narrowing the gap with leading closed-source systems.


\begin{table}[h!]
\centering
\renewcommand{\arraystretch}{1}
\setlength{\tabcolsep}{3.3pt}
\caption{\textbf{Quantitative evaluation results on GenEval}. The parameters of the generation component are denoted as \textit{\# Params}.}
\label{tab:geneval_results}
\begin{adjustbox}{width=\textwidth}
\begin{tabular}{lc|cccccc|c}
\toprule
\textbf{Model} & \textbf{\# Params} & \textbf{Single Object} & \textbf{Two Object} & \textbf{Counting} & \textbf{Colors} & \textbf{Position} & \textbf{Attribute Binding} & \textbf{Overall}$\uparrow$ \\
\midrule
\multicolumn{9}{c}{\textit{Closed-source Models}} \\
\midrule
GPT-Image-2~\cite{gpt_image_2}
& - & 0.99 & 0.98 & 0.85 & 0.93 & 0.85 & 0.77 & 0.89 \\

GPT-Image-1~\cite{GPT-Image-1}
& - & 0.99 & 0.92 & 0.85 & 0.92 & 0.75 & 0.61 & 0.84 \\

Seedream 4.0~\cite{seedream2025seedream}
& - & 0.99 & 0.92 & 0.72 & 0.91 & 0.76 & 0.74 & 0.84 \\

Nano-Banana-2.0~\cite{nanobanana2}
& - & 1.00 & 0.96 & 0.71 & 0.84 & 0.86 & 0.65 & 0.83 \\
\midrule
\multicolumn{9}{c}{\textit{Open-source Models}} \\
\midrule
\textbf{SenseNova-U1.5} & \textbf{8B} & 1.00 & 0.97 & 0.93 & 0.92 & 0.92 & 0.81 & \textbf{0.92} \\
SenseNova-U1~\cite{diao2026sensenovau1} & 8B & 1.00 & 0.96 & 0.92 & 0.92 & 0.91 & 0.76 & 0.91 \\
HiDream-O1-Image~\cite{cai2026hidream} & 8B & 1.00 & 0.99 & 0.79 & 0.89 & 0.93 & 0.78 & 0.90 \\
Tuna~\cite{liu2025tuna} & 7B & 1.00 & 0.97 & 0.81 & 0.91 & 0.88 & 0.83 & 0.90 \\
OneCAT~\cite{li2025onecat} & 9BA3B & 1.00 & 0.96 & 0.84 & 0.94 & 0.84 & 0.80 & 0.90 \\
NEO-unify~\cite{sensenova2026neounify} & 8B & 1.00 & 0.96 & 0.90 & 0.91 & 0.91 & 0.77 & 0.90 \\
Mogao~\cite{liao2025mogao} & 7B & 1.00 & 0.97 & 0.83 & 0.93 & 0.84 & 0.80 & 0.89 \\
ERNIE-Image~\cite{liu2026ernie}& 8B & 1.00 & 0.96 & 0.78 & 0.93 & 0.86 & 0.79 & 0.89 \\
Lumina-DiMOO~\cite{xin2025lumina} & 8B & 1.00 & 0.94 & 0.85 & 0.89 & 0.85 & 0.76 & 0.88 \\
Qwen-Image~\cite{wu2025qwenimagetechnicalreport} & 20B & 0.99 & 0.92 & 0.89 & 0.88 & 0.76 & 0.77 & 0.87 \\
NEO-unify~\cite{sensenova2026neounify} & 2B & 0.99 & 0.92 & 0.89 & 0.86 & 0.77 & 0.76 & 0.87 \\
Tuna-2~\cite{tuna2} & 7B & 0.99 & 0.96 & 0.80 & 0.91 & 0.84 & 0.76 & 0.87 \\
LLaDA-Image~\cite{chen2026llada}
& 6B & 1.00 & 0.98 & 0.53 & 0.93 & 0.78 & 0.84 & 0.85 \\
Boogu-Image 0.1 Base~\cite{chen2026boogu} & 10B & 0.99 & 0.95 & 0.80 & 0.84 & 0.85 & 0.68 & 0.85 \\
LongCat-Next~\cite{team2026longcat} & 68BA3B & - & - & - & - & - & - & 0.84 \\

\bottomrule
\end{tabular}
\end{adjustbox}
\end{table}

\begin{table}[h!]
\centering
\renewcommand{\arraystretch}{0.95}
\setlength{\tabcolsep}{3.3pt}
\small

\caption{\textbf{Quantitative evaluation results on GenEval2}.
The parameters of the generation component are denoted as \# Params.}
\label{tab:geneval2-results}

\begin{adjustbox}{max width=0.95\textwidth}
\begin{tabular}{lc|ccccc|c}
\toprule
\textbf{Model}
& \textbf{\# Params}
& \textbf{Attribute}
& \textbf{Counting}
& \textbf{Object}
& \textbf{Position}
& \textbf{Verb}
& \textbf{Overall}$\uparrow$ \\
\midrule

\multicolumn{8}{c}{\textit{Closed-source Models}} \\
\midrule

GPT-Image-2~\cite{gpt_image_2}
& --
& 0.98
& 0.89
& 0.98
& 0.90
& 0.70
& 0.81 \\

Qwen-Image-3.0~\citep{qwen_image_3}
& --
& 0.97
& 0.85
& 0.98
& 0.91
& 0.54
& 0.76 \\

Seedream-5.0-Pro~\cite{Seedream5}
& --
& 0.95
& 0.81
& 0.98
& 0.88
& 0.55
& 0.68 \\

Gemini-3-Pro-Image~\cite{deepmind_gemini3proimage_2025}
& --
& 0.97
& 0.74
& 0.96
& 0.84
& 0.47
& 0.59 \\

Qwen-Image-2.0~\cite{zhao2026qwenimage20technicalreport}
& --
& 0.93
& 0.70
& 0.98
& 0.66
& 0.47
& 0.48 \\

\midrule
\multicolumn{8}{c}{\textit{Open-source Models}} \\
\midrule

\textbf{SenseNova-U1.5 (w/ PE)}
& \textbf{8B}
& 0.99
& 0.83
& 0.98
& 0.90
& 0.55
& \textbf{0.71} \\

\textbf{SenseNova-U1.5}
& \textbf{8B}
& 0.89
& 0.77
& 0.98
& 0.77
& 0.41
& \textbf{0.55} \\

SenseNova-U1~\cite{diao2026sensenovau1}
& 8B
& 0.85
& 0.71
& 0.98
& 0.78
& 0.38
& 0.43 \\

Qwen-Image~\cite{wu2025qwenimagetechnicalreport}
& 20B
& -
& -
& -
& -
& -
& 0.42 \\

Z-Image~\cite{cai2025z}
& 6B
& -
& -
& -
& -
& -
& 0.31 \\

\bottomrule
\end{tabular}
\end{adjustbox}
\end{table}

\paragraph{GenEval.}
In Table~\ref{tab:geneval_results}, SenseNova-U1.5 achieves the best overall performance among open-source models on GenEval, reaching 0.92 and surpassing both SenseNova-U1 and larger baselines such as Qwen-Image. The improvements are particularly clear in counting, position, and attribute binding, while performance on single- and two-object generation remains consistently strong. These results indicate that U1.5 not only preserves basic object-level generation quality, but also improves compositional consistency and the ability to satisfy multiple visual constraints within a single prompt.

\paragraph{GenEval2.}

In Table~\ref{tab:geneval2-results}, SenseNova-U1.5 further shows stronger compositional reasoning and instruction following under complex constraints. With prompt enhancement, it substantially outperforms all evaluated open-source baselines, while further narrowing the gap to leading closed-source systems. The improvements are particularly notable in attribute, counting, and verb-related generation, indicating better modeling of object properties, quantities, and interactions. Notably, SenseNova-U1.5 is more capable of faithfully translating structured, relation-intensive prompts into coherent visual compositions, especially when multiple constraints need to be satisfied simultaneously.

\begin{table}[t]
\centering
\caption{\textbf{Quantitative evaluation results on OneIG-EN}.
The parameters of the generation component are denoted as \textit{\# Params}.}
\label{tab:oneig_en_results}
\renewcommand{\arraystretch}{1}
\setlength{\tabcolsep}{5pt}
\begin{adjustbox}{width=0.88\textwidth}
\begin{tabular}{lc|ccccc|c}
\toprule
\textbf{Model} & \textbf{\# Params} & \textbf{Alignment} & \textbf{Text} & \textbf{Reasoning} & \textbf{Style} & \textbf{Diversity} & \textbf{Overall}$\uparrow$ \\
\midrule
\multicolumn{8}{c}{\textit{Closed-source Models}} \\
\midrule
Nano-Banana-Pro~\cite{deepmind_gemini3proimage_2025} & - & 0.889 & 0.962 & 0.333 & 0.483 & 0.247 & 0.583 \\
Qwen-Image-3.0~\citep{qwen_image_3} & - & 0.888 & 0.976 & 0.336 & 0.420 & 0.193 & 0.563 \\
Seedream-5.0-Pro~\cite{Seedream5} & - & 0.899 & 0.886 & 0.341 & 0.431 & 0.207 & 0.553 \\
GPT-Image-2~\cite{gpt_image_2} & - & 0.903 & 0.741 & 0.352 & 0.443 & 0.139 & 0.516 \\
\midrule
\multicolumn{8}{c}{\textit{Open-source Models}} \\
\midrule

Emu3.5~\cite{cui2025emu35nativemultimodalmodels} & 32B & 0.902 & 0.994 & 0.345 & 0.427 & 0.151 & 0.564 \\
ERNIE-Image~\cite{liu2026ernie}& 8B & 0.891 & 0.967 & 0.295 & 0.447 & 0.169 & 0.554 \\
\textbf{SenseNova-U1.5} & \textbf{8B} & 0.883 & 0.988 & 0.329 & 0.402 & 0.159 & \textbf{0.552} \\
SenseNova-U1~\cite{diao2026sensenovau1} & 8B & 0.882 & 0.969 & 0.330 & 0.396 & 0.166 & 0.549 \\
Qwen-Image~\cite{wu2025qwenimagetechnicalreport} & 20B & 0.882 & 0.891 & 0.306 & 0.418 & 0.197 & 0.539 \\
HiDream-I1-Full~\cite{cai2025hidream} & 17B & 0.829 & 0.707 & 0.317 & 0.347 & 0.186 & 0.477 \\
SD3.5 Large~\cite{esser2024scaling} & 8B & 0.809 & 0.629 & 0.294 & 0.353 & 0.225 & 0.462 \\
FLUX.1 [Dev]~\cite{flux2024} & 12B & 0.786 & 0.523 & 0.253 & 0.368 & 0.238 & 0.434 \\
BAGEL~\cite{deng2025bagel} & 7B & 0.769 & 0.244 & 0.173 & 0.367 & 0.251 & 0.361 \\
BLIP3-o~\cite{chen2025blip3o} & 1.4B & 0.711 & 0.013 & 0.223 & 0.361 & 0.229 & 0.307 \\
\bottomrule
\end{tabular}
\end{adjustbox}
\end{table}

\begin{table}[h!]
\centering
\caption{\textbf{Quantitative evaluation results on OneIG-ZH}.
The parameters of the generation component are denoted as \textit{\# Params}.}
\label{tab:oneig_zh_results}
\renewcommand{\arraystretch}{1}
\setlength{\tabcolsep}{5pt}
\begin{adjustbox}{width=0.88\textwidth}
\begin{tabular}{lc|ccccc|c}
\toprule
\textbf{Model} & \textbf{\# Params} & \textbf{Alignment} & \textbf{Text} & \textbf{Reasoning} & \textbf{Style} & \textbf{Diversity} & \textbf{Overall}$\uparrow$ \\
\midrule
\multicolumn{8}{c}{\textit{Closed-source Models}} \\
\midrule
Nano-Banana-Pro~\cite{deepmind_gemini3proimage_2025} & - & 0.856 & 0.968 & 0.313 & 0.470 & 0.239 & 0.569 \\
Qwen-Image-3.0~\citep{qwen_image_3} & - & 0.836 & 0.950 & 0.320 & 0.413 & 0.196 & 0.543 \\
Seedream-5.0-Pro~\cite{Seedream5} & - & 0.855 & 0.887 & 0.302 & 0.421 & 0.202 & 0.533 \\
GPT-Image-2~\cite{gpt_image_2} & - & 0.861 & 0.868 & 0.321 & 0.432 & 0.146 & 0.526 \\
\midrule
\multicolumn{8}{c}{\textit{Open-source Models}} \\
\midrule
Qwen-Image~\cite{wu2025qwenimagetechnicalreport} & 20B & 0.825 & 0.963 & 0.267 & 0.405 & 0.279 & 0.548 \\
\textbf{SenseNova-U1.5} & \textbf{8B} & 0.831 & 0.985 & 0.307 & 0.392 & 0.165 & \textbf{0.536} \\
SenseNova-U1~\cite{diao2026sensenovau1} & 8B & 0.826 & 0.977 & 0.303 & 0.392 & 0.176 & 0.535 \\
Emu3.5~\cite{cui2025emu35nativemultimodalmodels} & 32B & 0.853 & 0.941 & 0.300 & 0.386 & 0.166 & 0.529 \\
ERNIE-Image~\cite{liu2026ernie}& 8B & 0.842 & 0.898 & 0.266 & 0.421 & 0.177 & 0.521 \\
BAGEL~\cite{deng2025bagel} & 7B & 0.672 & 0.365 & 0.186 & 0.357 & 0.268 & 0.370 \\
HiDream-I1-Full~\cite{cai2025hidream} & 17B & 0.620 & 0.205 & 0.256 & 0.304 & 0.300 & 0.337 \\
BLIP3-o~\cite{chen2025blip3o} & 1.4B & 0.608 & 0.092 & 0.213 & 0.369 & 0.233 & 0.303 \\
\bottomrule
\end{tabular}
\end{adjustbox}
\end{table}

\paragraph{OneIG-Bench.}
In Table~\ref{tab:oneig_en_results} and \ref{tab:oneig_zh_results}, SenseNova-U1.5 exhibits a notably balanced generation profile across English and Chinese evaluations. Note that its strongest advantage lies in bilingual text-centric generation. Besides, the model maintains competitive alignment, reasoning, and style scores despite its relatively compact 8B scale. This balanced performance is particularly important for practical generation scenarios, where accurate typography, semantic consistency, and visual quality must be satisfied simultaneously rather than optimized in isolation.

\begin{table}[t]
\centering
\caption{\textbf{Quantitative evaluation results on DPG-Bench}.
The parameters of the generation component are denoted as \textit{\# Params}.}
\label{tab:dpg_results}
\renewcommand{\arraystretch}{1}
\setlength{\tabcolsep}{3.1pt}
\begin{adjustbox}{width=0.75\textwidth}
\begin{tabular}{lc|ccccc|c}
\toprule
\textbf{Model} & \textbf{\# Params} & \textbf{Global} & \textbf{Entity} & \textbf{Attribute} & \textbf{Relation} & \textbf{Other} & \textbf{Overall}$\uparrow$ \\
\midrule
\multicolumn{8}{c}{\textit{Closed-source Models}} \\
\midrule
Seedream 4.5~\cite{seedream45} & - & 89.24 & 94.30 & 92.14 & 92.23 & 93.83 & 88.63 \\
GPT-Image-2~\cite{gpt_image_2} & - & 84.67 & 92.76 & 89.49 & 95.42 & 84.34 & 87.32 \\
Seedream-5.0-Pro~\cite{Seedream5} & - & 85.00 & 92.85 & 90.09 & 95.02 & 86.80 & 87.02 \\
Nano-Banana-Pro~\cite{deepmind_gemini3proimage_2025} & - & 84.19 & 92.47 & 89.58 & 94.74 & 81.12 & 86.59 \\
Qwen-Image-3.0~\citep{qwen_image_3} & - & 84.66 & 92.06 & 89.82 & 94.46 & 83.74 & 86.00 \\
\midrule
\multicolumn{8}{c}{\textit{Open-source Models}} \\
\midrule
Qwen-Image~\cite{wu2025qwenimagetechnicalreport} & 20B & 91.32 & 91.56 & 92.02 & 94.31 & 92.73 & 88.32 \\
Z-Image~\cite{cai2025z} & 6B & 93.39 & 91.22 & 93.16 & 92.22 & 91.52 & 88.14 \\
\textbf{SenseNova-U1.5} & \textbf{8B} & 86.54 & 93.61 & 91.52 & 93.33 & 92.13 & \textbf{88.11} \\
JoyAI-Image~\cite{joyai_image} & 16B & - & - & - & - & - & 88.05 \\
SenseNova-U1~\cite{diao2026sensenovau1} & 8B & 88.74 & 90.90 & 92.43 & 92.43 & 92.50  & 87.78 \\
LLaDA-Image~\cite{chen2026llada}
& 6B & 90.86 & 92.31 & 92.37 & 92.55 & 91.96  & 87.48\\ 
Boogu-Image-0.1-Base~\cite{chen2026boogu} & 10B & 89.33 & 90.64 & 92.22 & 93.72 & 93.32  & 87.13 \\
NEO-unify~\cite{sensenova2026neounify} & 8B & 91.00 & 91.53 & 92.06 & 94.14 & 90.43 & 86.71 \\
Tuna-2~\cite{tuna2} & 7B & 89.50 & 91.40 & 92.07 & 91.91 & 88.81 & 86.54 \\
NEO-unify~\cite{sensenova2026neounify} & 2B & 89.49 & 92.87 & 91.26 & 92.29 & 92.13 & 86.54 \\
BAGEL~\cite{deng2025bagel} & 7B & 88.94 & 90.37 & 91.29 & 90.82 & 88.67 & 85.07 \\
LongCat-Next~\cite{team2026longcat} & 68BA3B & - & - & - & - & - & 84.66 \\
OneCAT~\cite{li2025onecat} & 9BA3B & - & - & - & - & - & 84.53 \\
Mogao~\cite{liao2025mogao} & 7B & - & - & - & - & - & 84.33 \\
\bottomrule
\end{tabular}
\end{adjustbox}
\end{table}

\begin{table}[!h]
\centering
\caption{\textbf{Quantitative evaluation results on CVTG-2K}.
The parameters of the generation component are denoted as \textit{\# Params}.}
\label{tab:cvtg_results}
\renewcommand{\arraystretch}{1}
\setlength{\tabcolsep}{3pt}
\begin{adjustbox}{width=0.9\textwidth}
\begin{tabular}{lc|cc|cccc|c}
\toprule
\multirow{2}{*}{\textbf{Model}} & \multirow{2}{*}{\textbf{\# Params}} & \multirow{2}{*}{\textbf{NED}} & \multirow{2}{*}{\textbf{CLIPScore}} & \multicolumn{4}{c|}{\textbf{Word Accuracy}} & \multirow{2}{*}{\textbf{Average}$\uparrow$} \\
\cmidrule(lr){5-8}
& & & & 2 regions& 3 regions & 4 regions & 5 regions & \\
\midrule
\multicolumn{9}{c}{\textit{Closed-source Models}} \\
\midrule
Seedream 4.5~\cite{seedream45} & - & 0.948 & 0.807 & 0.878 & 0.895 & 0.908 & 0.901 & 0.899 \\
GPT-Image-1~\cite{GPT-Image-1} & - & 0.948 & 0.798 & 0.878 & 0.866 & 0.873 & 0.822 & 0.857 \\
Nano-Banana-Pro~\cite{deepmind_gemini3proimage_2025} & - & 0.875 & 0.737 & 0.737 & 0.775 & 0.786 & 0.793 & 0.779 \\
\midrule
\multicolumn{9}{c}{\textit{Open-source Models}} \\
\midrule
\textbf{SenseNova-U1.5} & \textbf{8B} & 0.977 & 0.819 & 0.933 & 0.941 & 0.955 & 0.954 & \textbf{0.948} \\
SenseNova-U1~\cite{diao2026sensenovau1} & 8B & 0.972 & 0.825 & 0.945 & 0.954 & 0.944 & 0.936 & 0.940 \\
HiDream-O1-Image~\cite{cai2026hidream}& 8B & 0.956 & 0.808 & 0.909 & 0.916 & 0.922 & 0.901 & 0.913 \\
Emu3.5~\cite{cui2025emu35nativemultimodalmodels} & 32B & - & - & - & - & - & - & 0.912 \\
LLaDA-Image~\cite{chen2026llada}
& 6B & 0.945 & 0.818 & 0.892 & 0.878 & 0.882 & 0.857 & 0.875 \\
JoyAI-Image~\cite{joyai_image} & 16B & 0.937 & 0.799 & - & - & - & - & 0.874 \\
Z-Image~\cite{cai2025z} & 6B & 0.937 & 0.797 & 0.901 & 0.872 & 0.865 & 0.851 & 0.867 \\
Boogu-Image-0.1-Base~\cite{chen2026boogu}& 10B & 0.927 & 0.801 & 0.852 & 0.863 & 0.866 & 0.863 & 0.863 \\
Qwen-Image~\cite{wu2025qwenimagetechnicalreport} & 20B & 0.912 & 0.802 & 0.837 & 0.836 & 0.831 & 0.816 & 0.829 \\
\bottomrule
\end{tabular}
\end{adjustbox}
\end{table}

\paragraph{DPG-Bench.}
In Table~\ref{tab:dpg_results}, SenseNova-U1.5 achieves an overall score of 88.11, placing it among the top-performing open-source models. It delivers consistently strong results across entity, attribute, relation, and other fine-grained dimensions, reflecting solid semantic grounding and compositional fidelity. Compared with SenseNova-U1, U1.5 improves the overall score while maintaining a balanced performance profile across different prompt components, indicating more reliable generation under complex and structured instructions.

\textbf{Text-centric Generation.}  We further examine text rendering under more demanding text-centric settings using CVTG-2K~\cite{du2025textcrafter} and LongText-Bench~\cite{geng2025x}, focusing on multi-region layouts and long-form bilingual text generation.

\paragraph{CVTG-2K.}
In Table~\ref{tab:cvtg_results}, SenseNova-U1.5 achieves the best overall performance among all evaluated models, reaching an average score of 0.948. Impressively, it demonstrates particularly strong gains in dense text rendering, with word accuracy remaining above 0.95 even in the more challenging four- and five-region settings. Compared with SenseNova-U1, U1.5 further improves NED and high-region word accuracy, while maintaining a comparable CLIPScore, indicating stronger text fidelity and robustness as the amount of rendered text increases.

\begin{table}[!h]
\centering
\caption{\textbf{Quantitative evaluation results on LongText-Bench}.
\textit{A} in \textit{\# Params} denotes activated parameters during inference.}
\label{tab:longtext_results}
\renewcommand{\arraystretch}{1}
\setlength{\tabcolsep}{3pt}
\begin{adjustbox}{width=0.70\textwidth}
\begin{tabular}{lc|cc}
\toprule
\textbf{Model} & \textbf{\# Params} & \textbf{LongText-Bench-EN}$\uparrow$ & \textbf{LongText-Bench-ZH}$\uparrow$ \\
\midrule
\multicolumn{4}{c}{\textit{Closed-source Models}} \\
\midrule
Seedream 4.5~\cite{seedream45} & - & 0.989 & 0.987 \\
Qwen-Image-3.0~\citep{qwen_image_3} & - & 0.982 & 0.971 \\
Nano-Banana-Pro~\cite{deepmind_gemini3proimage_2025} & - & 0.976 & 0.949 \\
GPT-Image-2~\cite{gpt_image_2} & - & 0.964 & 0.983 \\
Seedream-5.0-Pro~\cite{Seedream5} & - & 0.941 & 0.946 \\
\midrule
\multicolumn{4}{c}{\textit{Open-source Models}} \\
\midrule
\textbf{SenseNova-U1.5} & \textbf{8B} & \textbf{0.988} & \textbf{0.989} \\
SenseNova-U1~\cite{diao2026sensenovau1} & 8B & 0.979 & 0.962 \\
Emu3.5~\cite{cui2025emu35nativemultimodalmodels} & 32B & 0.976 & 0.928 \\
JoyAI-Image~\cite{joyai_image} & 16B & 0.963 & 0.963 \\
Boogu-Image-0.1-Base~\cite{chen2026boogu}  & 10B & 0.952 & 0.969 \\
ERNIE-Image~\cite{liu2026ernie}& 8B  & 0.968 & 0.959 \\
Qwen-Image~\cite{wu2025qwenimagetechnicalreport} & 20B & 0.943 & 0.946 \\
LLaDA-Image~\cite{chen2026llada}
& 6B & 0.923 & 0.913 \\
Z-Image~\cite{cai2025z} & 6B & 0.935 & 0.936 \\
LongCat-Next~\cite{team2026longcat} & 68BA3B & 0.932 & 0.891 \\
X-Omni~\cite{geng2025x} & 12B & 0.900 & 0.814 \\
InternVL-U~\cite{tian2026internvludemocratizingunifiedmultimodal} & 1.7B & 0.738 & 0.860 \\
NEO-unify~\cite{sensenova2026neounify} & 2B & 0.748 & 0.495 \\
\bottomrule
\end{tabular}
\end{adjustbox}
\end{table}

\begin{table*}[!h]
\centering
\caption{\textbf{Quantitative evaluation results on IGenBench}. 
The parameters of the generation component are denoted as \textit{\# Params}.}
\label{tab:igenbench_results}
\small
\setlength{\tabcolsep}{2pt}
\renewcommand{\arraystretch}{1}
\begin{adjustbox}{max width=\textwidth}
\begin{tabular}{l c|cccccccccc|cc}
\toprule
\multirow{2}{*}{\textbf{Model}} &
\multirow{2}{*}{\textbf{\# Params}} &
\multicolumn{10}{c|}{\textbf{Question Type}} &
\multicolumn{2}{c}{\textbf{Overall}} \\
\cmidrule(lr){3-12} \cmidrule(lr){13-14}
 & & \scriptsize\faCalculator~Comp.
 & \scriptsize\faCode~Enc.
 & \scriptsize\faChartLine~Order
 & \scriptsize\faMarker~Marks
 & \scriptsize\faEdit~Anno.
 & \scriptsize\faRuler~Axes
 & \scriptsize\faListUl~Leg.
 & \scriptsize\faChartPie~Chart
 & \scriptsize\faHeading~Title
	 & \scriptsize\faGem~Deco.
	 & Q-ACC$\uparrow$ & I-ACC \\
\midrule
\multicolumn{14}{c}{\textit{Closed-source Models}} \\
\midrule
Nano-Banana-Pro~\cite{deepmind_gemini3proimage_2025}
& -
& 0.84
& 0.86
& 0.90
& 0.87
& 0.93
& 0.93
& 0.96
& 0.92
& 0.98
& 0.94
& 0.90
& 0.49\\
GPT-Image-2~\cite{gpt_image_2}
& -
& 0.89 & 0.61 & 0.93 & 0.87 & 0.94 & 0.98 & 0.95 & 0.93 & 0.99 & 0.95
& 0.88 & 0.38 \\
Qwen-Image-3.0~\citep{qwen_image_3}
& -
& 0.65 & 0.43 & 0.77 & 0.77 & 0.84 & 0.84 & 0.87 & 0.85 & 0.80 & 0.88
& 0.74 & 0.13 \\
Seedream-5.0-Pro~\cite{Seedream5}
& -
& 0.64 & 0.47 & 0.71 & 0.73 & 0.77 & 0.85 & 0.83 & 0.83 & 0.86 & 0.88
& 0.73 & 0.17 \\
\midrule
\multicolumn{14}{c}{\textit{Open-source Models}} \\
\midrule
\textbf{SenseNova-U1.5 (w/ PE)}
& \textbf{8B}
& 0.65 & 0.47 & 0.76 & 0.77 & 0.82
& 0.82 & 0.89 & 0.84 & 0.87 & 0.92
& \textbf{0.76} & \textbf{0.17} \\
\textbf{SenseNova-U1.5}
& \textbf{8B}
& 0.41
& 0.33
& 0.58
& 0.56
& 0.66
& 0.67
& 0.83
& 0.72
& 0.83
& 0.88
& \textbf{0.62}
& \textbf{0.08} \\
SenseNova-U1~\cite{diao2026sensenovau1}
& 8B
& 0.27 & 0.23 & 0.49 & 0.45 & 0.54 & 0.61 & 0.70 & 0.65 & 0.74 & 0.82
& 0.51 & 0.04 \\
Qwen-Image~\cite{wu2025qwenimagetechnicalreport}
& 20B
& 0.10 & 0.13 & 0.19 & 0.29 & 0.43 & 0.37 & 0.51 & 0.48 & 0.56 & 0.78
& 0.36 & 0.01 \\
Z-Image-Turbo~\cite{cai2025z}
& 6B
& 0.10 & 0.16 & 0.16 & 0.25 & 0.38 & 0.31 & 0.58 & 0.42 & 0.61 & 0.73
& 0.35 & 0.00 \\
HiDream-I1~\cite{cai2025hidream}
& 17B
& 0.01 & 0.03 & 0.03 & 0.10 & 0.07 & 0.14 & 0.10 & 0.26 & 0.19 & 0.20
& 0.11 & 0.00 \\
FLUX.1-dev~\cite{flux2024}
& 12B
& 0.00 & 0.03 & 0.01 & 0.08 & 0.06 & 0.06 & 0.01 & 0.24 & 0.09 & 0.39
& 0.10 & 0.00 \\
\bottomrule
\end{tabular}
\end{adjustbox}
\end{table*}

\paragraph{LongText-Bench.}
In Table~\ref{tab:longtext_results}, SenseNova-U1.5 achieves leading open-source performance on both English and Chinese tracks, with particularly strong gains over SenseNova-U1 in Chinese. Its accuracy remains robust as text length and layout complexity increase, demonstrating improved stability in dense, structured, and bilingual text rendering.

\textbf{Complex Infographic Generation.}   
We further evaluate SenseNova-U1.5 on complex infographic and commercial content generation using IGenBench~\cite{tang2026igenbench} and BizGenEval~\cite{li2026bizgeneval}. They require the model to jointly satisfy textual content, layout organization, chart semantics, visual attributes, and domain knowledge.

\paragraph{IGenBench.}
As shown in Table~\ref{tab:igenbench_results}, SenseNova-U1.5 already achieves the strongest open-source performance without prompt enhancement, and improves further with PE. These gains indicate more reliable infographic generation and stronger structured visual planning, although a gap to the strongest proprietary systems remains.

\begin{table}[t]
\centering
\small
\caption{\textbf{Quantitative evaluation results on BizGenEval}. Each cell reports hard / easy testset scores.}
\label{tab:bizgeneval_results}
\resizebox{0.8\linewidth}{!}{%
\begin{tabular}{lc|cccc|c}
\toprule
\textbf{Model} & \textbf{\# Params} & \textbf{Layout} & \textbf{Attribute} & \textbf{Text} & \textbf{Knowledge} & \textbf{Average}$\uparrow$ \\
\midrule
\multicolumn{7}{c}{\textit{Closed-source Models}} \\
\midrule
GPT-Image-2~\cite{gpt_image_2} & - & 88.5 / 95.3 &  81.9 / 91.0 &  83.9 / 92.6  &  74.2 / 90.9 & 82.1 / 92.5 \\
Nano-Banana-Pro~\cite{deepmind_gemini3proimage_2025} & - & 72.2 / 91.2 & 65.6 / 92.2 & 86.4 / 95.0 & 82.6 / 96.2 & 76.7 / 93.7 \\
Qwen-Image-3.0~\citep{qwen_image_3} & - & 79.3 / 89.5 & 74.0 / 87.6 & 72.0 / 88.3 & 76.8 / 90.5 & 75.5 / 89.0 \\
Nano-Banana-2.0~\cite{nanobanana2} & - & 68.4 / 91.0 & 57.4 / 91.6 & 83.4 / 94.6 & 64.6 / 93.0 & 68.5 / 92.5 \\
Seedream-5.0-Pro~\cite{Seedream5} & - & 75.5 / 87.6 & 74.1 / 87.3 & 63.2 / 81.1 & 49.8 / 75.3 & 65.7 / 82.8 \\
\midrule
\multicolumn{7}{c}{\textit{Open-source Models}} \\
\midrule
\textbf{SenseNova-U1.5 (w/ PE)}
& \textbf{8B}
& 82.5 / 92.3
& 66.5 / 86.9
& 67.4 / 86.5
& 72.0 / 87.9
& \textbf{72.2 / 88.4} \\

\textbf{SenseNova-U1.5}
& \textbf{8B}
& 78.1 / 91.2
& 55.9 / 80.0
& 66.6 / 86.1
& 4.8 / 23.4
& \textbf{51.4 / 70.2} \\
SenseNova-U1~\cite{diao2026sensenovau1} & 8B & 61.6 / 81.6 & 47.5 / 72.8 & 46.3 / 74.6 & 3.5 / 17.9 & 39.7 / 61.7 \\
Emu3.5~\cite{cui2025emu35nativemultimodalmodels} & 32B & 30.4 / 63.4 & 14.2 / 52.6 & 7.0 / 33.6 & 1.2 / 11.0 & 13.2 / 40.2 \\
HunyuanImage-3.0~\cite{cao2025hunyuanimage} & 80BA13B & 27.8 / 65.0 & 13.8 / 53.6 & 10.2 / 39.6 & 0.0 / 2.0 & 13.0 / 40.1 \\
Z-Image~\cite{cai2025z} & 6B & 26.8 / 69.2 & 2.6 / 47.6 & 2.8 / 45.0 & 0.6 / 13.2 & 8.2 / 43.8 \\
Qwen-Image-2512~\cite{wu2025qwenimagetechnicalreport} & 20B & 22.2 / 70.6 & 1.2 / 47.8 & 1.8 / 39.2 & 0.0 / 6.4 & 6.3 / 41.0 \\
FLUX.2-dev~\cite{flux-2-2025} & 32B & 17.2 / 67.8 & 1.2 / 49.2 & 1.0 / 43.0 & 0.0 / 8.2 & 4.9 / 42.0 \\
Qwen-Image~\cite{wu2025qwenimagetechnicalreport} & 20B & 10.4 / 51.2 & 0.2 / 22.2 & 0.6 / 17.6 & 0.0 / 4.4 & 2.8 / 23.8 \\
\bottomrule
\end{tabular}
}
\end{table}

\begin{table}[h!]
\centering
\caption{\textbf{Quantitative evaluation results on WISE}~\cite{niu2025wise}.
The parameters of the generation component are denoted as \textit{\# Params}.
}
\label{tab:wise_results}
\renewcommand{\arraystretch}{1}
\setlength{\tabcolsep}{3pt}
\begin{adjustbox}{width=0.84\textwidth}
\begin{tabular}{lc|cccccc|c}
\toprule
\textbf{Model} & \textbf{\# Params} & \textbf{Cultural} & \textbf{Time} & \textbf{Space} & \textbf{Biology} & \textbf{Physics} & \textbf{Chemistry} & \textbf{Overall}$\uparrow$ \\
\midrule
\multicolumn{9}{c}{\textit{Closed-source Models}} \\
\midrule
Nano-Banana-Pro~\cite{deepmind_gemini3proimage_2025} & -  &0.89& 0.80& 0.89& 0.88 &0.86 &0.85& 0.87\\
GPT-Image-1~\cite{GPT-Image-1} & - & 0.81 & 0.71 & 0.89 & 0.83 & 0.79 & 0.74 & 0.80 \\
Seedream 4.0~\cite{seedream2025seedream}&- &0.78& 0.73& 0.85 &0.79& 0.84& 0.67& 0.78 \\
\midrule
\multicolumn{9}{c}{\textit{Open-source Models}} \\
\midrule
\textbf{SenseNova-U1.5 (w/ CoT)} & \textbf{8B} & 0.82 & 0.73 & 0.86 & 0.80 & 0.86 & 0.76 & \textbf{0.81} \\
SenseNova-U1-SFT (w/ CoT)~\cite{diao2026sensenovau1} & 8B & 0.78 & 0.73 & 0.82 & 0.80 & 0.85 & 0.77 & 0.78 \\
NEO-unify (w/ CoT)~\cite{sensenova2026neounify} & 8B & 0.73 & 0.67 & 0.79 & 0.70 & 0.75 & 0.66 & 0.72 \\
\textbf{SenseNova-U1.5} & \textbf{8B} & 0.66 & 0.68 & 0.82 & 0.69 & 0.80 & 0.66 & \textbf{0.70} \\
BAGEL (w/ CoT)~\cite{deng2025bagel} & 7B & 0.76 & 0.69 & 0.75 & 0.65  &0.75  &0.58  &0.70 \\
SenseNova-U1-SFT~\cite{diao2026sensenovau1} & 8B & 0.65 & 0.66 & 0.82 & 0.68 & 0.81 & 0.66 & 0.69 \\
Qwen-Image~\cite{wu2025qwenimagetechnicalreport} & 20B & 0.63 & 0.62 & 0.76 & 0.60 & 0.72 & 0.39 & 0.63 \\
BLIP3-o~\cite{chen2025blip3o} & 1.4B & - & - & - & - & - & - & 0.62 \\
Emu3.5~\cite{cui2025emu35nativemultimodalmodels} & 32B & - & - & - & - & - & - & 0.58 \\
LongCat-Next~\cite{team2026longcat} & 68BA3B & - & - & - & - & - & - & 0.57 \\
UniWorld-V1~\cite{lin2025uniworld} & 12B & 0.53 & 0.55 & 0.73 & 0.45 & 0.59 & 0.41 & 0.55 \\
BAGEL~\cite{deng2025bagel} & 7B & 0.44 & 0.52 & 0.65 & 0.42 & 0.62 & 0.41 & 0.49 \\
NEO-unify~\cite{sensenova2026neounify} & 8B & - & - & - & - & - & - & 0.47 \\
\bottomrule
\end{tabular}
\end{adjustbox}
\end{table}

\paragraph{BizGenEval.}
As shown in Table~\ref{tab:bizgeneval_results}, SenseNova-U1.5 substantially improves over SenseNova-U1 on both easy and hard splits, with prompt enhancement further achieving leading open-source performance. The gains are especially clear in knowledge-intensive cases, while remaining consistent across layout, attribute, and text-related dimensions.

\textbf{Reasoning-centric Generation.}
We further evaluate knowledge- and reasoning-intensive generation capabilities using WISE~\cite{niu2025wise}, covering cultural knowledge, time, space, biology, physics, and chemistry.

\paragraph{WISE.}
As shown in Table~\ref{tab:wise_results}, SenseNova-U1.5 is already competitive without chain-of-thought (CoT) reasoning, while enabling CoT yields a substantial improvement and establishes leading open-source performance. The gains are particularly evident in knowledge-intensive domains such as cultural knowledge, biology, and chemistry, highlighting the benefit of explicit reasoning for grounding visual generation in world knowledge.

\begin{table}[t]
\centering
\caption{\textbf{Quantitative evaluation results on ImgEdit}.
The parameters of the generation component are denoted as \textit{\# Params}.
}
\label{tab:u15_imgedit_results}
\renewcommand{\arraystretch}{1}
\setlength{\tabcolsep}{2pt}
\begin{adjustbox}{width=\textwidth}
\begin{tabular}{lc|ccccccccc|c}
\toprule
\textbf{Model} & \textbf{\# Params} & \textbf{Add} & \textbf{Adjust} & \textbf{Extract} & \textbf{Replace} & \textbf{Remove} & \textbf{Background} & \textbf{Style} & \textbf{Hybrid} & \textbf{Action} & \textbf{Overall}$\uparrow$ \\
\midrule
\multicolumn{12}{c}{\textit{Closed-source Models}} \\
\midrule
UniWorld-V2~\cite{li2025uniworld} & - & 4.29 & 4.44 & 4.32 & 4.69 & 4.72 & 4.41 & 4.91 & 3.83 & 4.83 & 4.49 \\
Nano-Banana-Pro~\cite{deepmind_gemini3proimage_2025} & - & 4.44 & 4.62 & 3.42 & 4.60 & 4.63 & 4.32 & 4.97 & 3.64 & 4.69 & 4.37 \\
Seedream 4.5~\cite{seedream45} & - & 4.57 & 4.65 & 2.97 & 4.66 & 4.46 & 4.37 & 4.92 & 3.71 & 4.56 & 4.32 \\
Seedream 4.0~\cite{seedream2025seedream} & - & 4.33 & 4.38 & 3.89 & 4.65 & 4.57 & 4.35 & 4.22 & 3.71 & 4.61 & 4.30 \\
GPT-Image-1~\cite{GPT-Image-1} & - & 4.61 & 4.33 & 2.90 & 4.35 & 3.66 & 4.57 & 4.93 & 3.96 & 4.89 & 4.20 \\
\midrule
\multicolumn{12}{c}{\textit{Open-source Models}} \\
\midrule
\textbf{SenseNova-U1.5} & \textbf{8B} & 4.64 & 4.66 & 4.20 & 4.84 & 4.54 & 4.59 & 4.91 & 4.36 & 4.61 & \textbf{4.59} \\
FireRed-Image-Edit~\cite{firered2026rededit} & 20B & 4.55 & 4.66 & 4.34 & 4.75 & 4.58 & 4.45 & 4.97 & 4.07 & 4.71 & 4.56 \\
Qwen-Image-Edit-2511~\cite{wu2025qwenimagetechnicalreport} & 20B & 4.54 & 4.57 & 4.13 & 4.70 & 4.46 & 4.36 & 4.89 & 4.16 & 4.81 & 4.51 \\
Boogu-Image-0.1-Edit~\cite{chen2026boogu}& 10B & 4.71 & 4.50 & 3.69 & 4.65 & 4.75 & 4.44 & 4.94 & 4.04 & 4.90 & 4.51 \\
LongCat-Image-Edit~\cite{team2025longcat} & 6B & 4.44 & 4.53 & 3.83 & 4.80 & 4.60 & 4.33 & 4.92 & 3.75 & 4.82 & 4.45 \\
Emu3.5~\cite{cui2025emu35nativemultimodalmodels} & 32B & 4.61 & 4.32 & 3.96 & 4.84 & 4.58 & 4.35 & 4.79 & 3.69 & 4.57 & 4.41 \\
FLUX.2 [Dev]~\cite{flux-2-2025} & 32B & 4.50 & 4.18 & 3.83 & 4.65 & 4.65 & 4.31 & 4.88 & 3.46 & 4.70 & 4.35 \\
Z-Image-Edit~\cite{cai2025z} & 6B & 4.40 & 4.14 & 4.30 & 4.57 & 4.13 & 4.14 & 4.85 & 3.63 & 4.50 & 4.30 \\
Qwen-Image-Edit~\cite{wu2025qwenimagetechnicalreport} & 20B & 4.38 & 4.16 & 3.43 & 4.66 & 4.14 & 4.38 & 4.81 & 3.82 & 4.69 & 4.27 \\
SenseNova-U1~\cite{diao2026sensenovau1} & 8B & 3.83 & 4.15 & 3.12 & 4.32 & 3.26 & 4.18 & 4.85 & 3.03 & 4.41 & 3.90 \\
OmniGen2~\cite{wu2025omnigen2} & 4B & 3.57 & 3.06 & 1.77 & 3.74 & 3.20 & 3.57 & 4.81 & 2.52 & 4.68 & 3.44 \\
\bottomrule
\end{tabular}
\end{adjustbox}
\end{table}

  \begin{table}[!h]
  \centering
  \caption{\textbf{Quantitative evaluation results on GEdit-Bench}.
  The parameters of the generation component are denoted as \textit{\# Params}.}
  \label{tab:u15_gedit_results}
  \renewcommand{\arraystretch}{1}
  \setlength{\tabcolsep}{4pt}
  \begin{adjustbox}{width=0.7\textwidth}
  \begin{tabular}{lc|ccc|ccc}
  \toprule
  \multirow{2}{*}{\textbf{Model}} &
  \multirow{2}{*}{\textbf{\# Params}} &
  \multicolumn{3}{c|}{\textbf{GEdit-Bench-EN}} &
  \multicolumn{3}{c}{\textbf{GEdit-Bench-CN}} \\
  \cmidrule(lr){3-5} \cmidrule(lr){6-8}
  & & G\_SC & G\_PQ & G\_O $\uparrow$ & G\_SC & G\_PQ & G\_O $\uparrow$ \\
  \midrule

  \multicolumn{8}{c}{\textit{Closed-source Models}} \\
  \midrule
  GPT-Image-2~\cite{gpt_image_2} & - & 9.42 & 8.31 & 8.73 & 9.43 & 8.31 & 8.76 \\
  Seedream 5.0 Pro~\cite{Seedream5} & - & 9.02 & 8.45 & 8.53 & 9.07 & 8.37 & 8.53 \\
  Nano-Banana-Pro~\cite{deepmind_gemini3proimage_2025} & - & 7.86 & 8.33 & 7.54 & 7.51 & 8.31 & 7.25 \\
  Seedream 4.0~\cite{seedream2025seedream} & - & 8.24 & 8.08 & 7.68 & 8.19 & 8.14 & 7.71 \\
  Qwen-Image-2.0~\cite{zhao2026qwenimage20technicalreport} & - & 9.02 & 8.02 & 8.37 & 8.99 & 8.04 & 8.35 \\
  Qwen-Image-3.0~\citep{qwen_image_3} & - & 8.68 & 8.37 & 8.40 & 8.84 & 8.34 & 8.49 \\
  \midrule
  \multicolumn{8}{c}{\textit{Open-source Models}} \\
  \midrule
  \textbf{SenseNova-U1.5} & \textbf{8B} & 9.15 & 7.79 & \textbf{8.26} & 9.06 & 7.79 & \textbf{8.24} \\
  FireRed-Image-Edit~\cite{firered2026rededit} & 20B & 8.36 & 8.25 & 7.94 & 8.28 & 8.22 & 7.88 \\
  Qwen-Image-Edit-2511~\cite{wu2025qwenimagetechnicalreport} & 20B & 8.00 & 7.86 & 7.56 & 7.82 & 7.79 & 7.52 \\
  LongCat-Image-Edit~\cite{team2025longcat} & 6B & 8.18 & 8.00 & 7.64 & 8.08 & 7.99 & 7.60 \\
  Emu3.5~\cite{cui2025emu35nativemultimodalmodels} & 32B & 8.11 & 7.70 & 7.59 & - & - & - \\
  Z-Image-Edit~\cite{cai2025z} & 6B & 8.11 & 7.72 & 7.57 & 8.03 & 7.80 & 7.54 \\
  Qwen-Image-Edit [2509]~\cite{wu2025qwenimagetechnicalreport} & 20B & 8.15 & 7.86 & 7.54 & 8.05 & 7.88 & 7.49 \\
  SenseNova-U1~\cite{diao2026sensenovau1} & 8B & 8.27 & 7.49 & 7.47 & 8.61 & 7.36 & 7.70 \\
  FLUX.2 [Dev]~\cite{flux-2-2025} & 32B & 7.84 & 8.06 & 7.41 & 7.69 & 8.04 & 7.28 \\
  LLaDA-Image~\cite{chen2026llada} & 6B & 8.04 & 7.18 & 7.33 & 7.71 & 7.59 & 7.29 \\
  BAGEL~\cite{deng2025bagel} & 7B & 7.36 & 6.83 & 6.52 & 7.34 & 6.85 & 6.50 \\

  \bottomrule
  \end{tabular}
  \end{adjustbox}
  \end{table}

\subsection{Image Editing}

\textbf{General Editing.}
We evaluate general editing capability on ImgEdit~\cite{ye2025imgedit}, GEdit-Bench~\cite{liu2025step1x}, WeEdit~\cite{zhang2026weedit}, and OmniRef-Bench~\cite{huang2026scaling}, covering diverse edit types, semantic consistency, text-centric and reference-based editing.

\paragraph{ImgEdit.}

As shown in Table~\ref{tab:u15_imgedit_results}, SenseNova-U1.5 achieves leading overall performance across a broad range of edit types, with substantial gains over SenseNova-U1. The improvements are consistent across categories, indicating stronger edit execution while maintaining source-image preservation and visual quality.

\paragraph{GEdit-Bench.}
In Table~\ref{tab:u15_gedit_results}, SenseNova-U1.5 achieves the strongest overall performance. The gains are particularly evident in semantic consistency, indicating that the model can more reliably identify the intended edit, modify the correct visual content, and preserve the semantic context of the original image. At the same time, U1.5 maintains competitive perceptual quality, suggesting that improved instruction execution does not come at the cost of visual realism or local coherence. This reflects a better balance between edit correctness, preservation, and overall image quality.

\paragraph{WeEdit.}
\begin{table*}[t]
\centering
\caption{\textbf{Quantitative evaluation results on the WeEdit benchmark.}
Note that IA, TC, and BP denote Instruction Adherence, Text Clarity, and Background Preservation, respectively. Avg. is the arithmetic mean of the three Overall scores.}
\label{tab:u15_weedit_results}
\renewcommand{\arraystretch}{1}
\setlength{\tabcolsep}{1.3pt}
\begin{adjustbox}{width=\textwidth}
\begin{tabular}{l|ccc|ccc|ccc|ccc|ccc|ccc|ccc|ccc|ccc|c}
\toprule
\multirow{2}{*}{\textbf{Model}} & \multicolumn{3}{c|}{\textbf{Add}} & \multicolumn{3}{c|}{\textbf{Replace}} & \multicolumn{3}{c|}{\textbf{Delete}} & \multicolumn{3}{c|}{\textbf{Rearrange}} & \multicolumn{3}{c|}{\textbf{Translate}} & \multicolumn{3}{c|}{\textbf{Style}} & \multicolumn{3}{c|}{\textbf{Combined}} & \multicolumn{3}{c|}{\textbf{Reasoning}} & \multicolumn{3}{c|}{\textbf{Overall}} & \multirow{2}{*}{\textbf{Avg.}$\uparrow$} \\
\cmidrule(lr){2-4}\cmidrule(lr){5-7}\cmidrule(lr){8-10}\cmidrule(lr){11-13}\cmidrule(lr){14-16}\cmidrule(lr){17-19}\cmidrule(lr){20-22}\cmidrule(lr){23-25}\cmidrule(lr){26-28}
& \textbf{IA} & \textbf{TC} & \textbf{BP} & \textbf{IA} & \textbf{TC} & \textbf{BP} & \textbf{IA} & \textbf{TC} & \textbf{BP} & \textbf{IA} & \textbf{TC} & \textbf{BP} & \textbf{IA} & \textbf{TC} & \textbf{BP} & \textbf{IA} & \textbf{TC} & \textbf{BP} & \textbf{IA} & \textbf{TC} & \textbf{BP} & \textbf{IA} & \textbf{TC} & \textbf{BP} & \textbf{IA} & \textbf{TC} & \textbf{BP} & \\
\midrule
\multicolumn{29}{c}{\textit{Closed-source Models}} \\
\midrule
Gemini-3-Pro-Image~\cite{deepmind_gemini3proimage_2025} & 9.38 & 9.66 & 9.43 & 9.06 & 8.80 & 8.05 & 8.33 & 8.56 & 8.04 & 8.05 & 8.49 & 8.26 & 8.08 & 9.07 & 9.55 & 9.76 & 9.51 & 9.50 & 9.31 & 9.72 & 9.45 & 4.91 & 9.47 & 9.42 & 8.58 & 9.10 & 8.85 & 8.84 \\
GPT-Image-1.5~\cite{GPT-Image-1.5} & 8.02 & 9.32 & 6.80 & 7.16 & 7.58 & 5.55 & 6.43 & 6.49 & 3.82 & 5.44 & 6.91 & 5.35 & 4.09 & 6.97 & 7.71 & 9.30 & 9.44 & 8.31 & 8.12 & 9.12 & 6.34 & 3.37 & 7.02 & 7.93 & 6.52 & 7.78 & 6.15 & 6.82 \\
Seedream 4.5~\cite{seedream45} & 6.86 & 8.21 & 6.40 & 7.23 & 7.93 & 5.91 & 6.60 & 7.37 & 4.70 & 5.29 & 7.08 & 5.36 & 4.97 & 6.93 & 7.15 & 8.76 & 9.08 & 8.37 & 7.33 & 8.62 & 7.66 & 2.17 & 5.69 & 7.67 & 6.29 & 7.66 & 6.38 & 6.78 \\
Gemini-2.5-Flash-Image~\cite{google2025gemini25flashmodelcard} & 5.88 & 8.06 & 7.58 & 3.35 & 6.05 & 7.50 & 6.01 & 8.17 & 6.05 & 1.77 & 6.86 & 7.11 & 1.75 & 6.06 & 8.83 & 7.95 & 9.54 & 9.58 & 4.08 & 7.57 & 8.83 & 1.70 & 4.82 & 8.76 & 3.92 & 7.14 & 7.80 & 6.29 \\
\midrule
\multicolumn{29}{c}{\textit{Open-source Models}} \\
\midrule

\textbf{SenseNova-U1.5} & 8.40 & 8.54 & 8.52 & 8.83 & 8.56 & 8.22 & 8.20 & 8.75 & 7.38 & 5.71 & 7.18 & 7.67 & 2.19 & 4.16 & 8.11 & 8.60 & 8.58 & 7.91 & 8.58 & 8.52 & 8.68 & 1.84 & 4.17 & 7.94 & 6.81 & 7.50 & 8.08 & \textbf{7.46} \\
FireRed-Image-Edit~\cite{qiao2026fireredimageedit} & 5.44 & 7.54 & 7.67 & 4.58 & 6.30 & 6.26 & 6.80 & 8.81 & 5.97 & 1.76 & 4.78 & 5.52 & 1.39 & 2.67 & 8.44 & 9.30 & 9.23 & 9.78 & 4.19 & 7.23 & 7.48 & 1.25 & 5.45 & 8.90 & 4.15 & 6.33 & 7.14 & 5.87 \\
HY-image-3-instruct~\cite{cao2025hunyuanimage} & 6.04 & 7.55 & 7.68 & 3.91 & 5.42 & 5.79 & 6.32 & 7.85 & 4.69 & 1.73 & 4.26 & 6.59 & 1.45 & 3.52 & 8.67 & 9.43 & 9.35 & 9.52 & 4.73 & 6.62 & 7.43 & 1.12 & 4.67 & 8.65 & 4.16 & 5.99 & 7.03 & 5.73 \\
Qwen-Image-Edit-2509~\cite{wu2025qwenimagetechnicalreport} & 4.96 & 7.05 & 6.55 & 3.44 & 5.17 & 5.90 & 4.97 & 7.84 & 5.07 & 1.35 & 4.90 & 5.91 & 1.26 & 2.72 & 8.77 & 8.97 & 9.15 & 9.11 & 3.90 & 6.48 & 7.24 & 1.22 & 5.17 & 8.51 & 3.49 & 5.84 & 6.80 & 5.38 \\
FLUX.2-dev~\cite{flux-2-2025} & 5.40 & 7.59 & 4.92 & 3.59 & 6.09 & 4.95 & 3.07 & 8.14 & 5.27 & 1.44 & 5.73 & 5.84 & 1.24 & 3.59 & 8.78 & 8.04 & 8.88 & 8.31 & 4.60 & 7.31 & 5.52 & 1.26 & 6.33 & 9.55 & 3.37 & 6.53 & 6.19 & 5.36 \\
LongCat-Image-Edit~\cite{team2025longcat} & 5.23 & 6.10 & 6.80 & 3.03 & 4.51 & 5.16 & 4.65 & 7.18 & 5.84 & 1.33 & 4.91 & 5.92 & 1.29 & 4.42 & 8.29 & 8.40 & 8.95 & 9.27 & 3.89 & 5.47 & 6.87 & 1.17 & 5.05 & 8.38 & 3.39 & 5.59 & 6.71 & 5.23 \\
Step1X-Edit-v1.2~\cite{liu2025step1x} & 2.73 & 3.98 & 7.38 & 2.44 & 3.22 & 6.75 & 5.68 & 7.01 & 6.45 & 1.26 & 3.46 & 4.76 & 1.34 & 2.84 & 7.64 & 7.78 & 8.54 & 8.67 & 2.12 & 3.79 & 7.89 & 1.09 & 5.81 & 9.37 & 2.78 & 4.36 & 7.03 & 4.72 \\
Emu3.5~\cite{cui2025emu35nativemultimodalmodels} & 4.89 & 6.60 & 3.72 & 3.26 & 4.60 & 2.81 & 4.76 & 5.20 & 2.29 & 1.46 & 3.94 & 3.36 & 1.27 & 2.73 & 5.93 & 8.27 & 8.66 & 8.41 & 3.96 & 5.98 & 4.13 & 1.23 & 3.34 & 6.32 & 3.42 & 4.96 & 4.07 & 4.15 \\
Qwen-Image-Edit-2511~\cite{wu2025qwenimagetechnicalreport} & 4.80 & 5.94 & 3.73 & 3.36 & 3.73 & 3.77 & 3.89 & 4.46 & 3.34 & 1.42 & 2.67 & 3.95 & 1.28 & 1.60 & 6.97 & 8.50 & 8.88 & 8.48 & 3.27 & 4.21 & 4.23 & 1.09 & 1.97 & 6.04 & 3.18 & 3.93 & 4.63 & 3.91 \\
BAGEL~\cite{deng2025bagel} & 2.32 & 3.03 & 5.99 & 1.48 & 2.41 & 4.30 & 3.03 & 6.46 & 2.92 & 1.07 & 3.53 & 5.66 & 1.03 & 4.88 & 7.68 & 7.36 & 7.76 & 8.98 & 1.44 & 2.33 & 5.83 & 1.02 & 4.42 & 8.89 & 1.97 & 4.01 & 5.75 & 3.91 \\
OmniGen2~\cite{wu2025omnigen2} & 1.72 & 3.11 & 3.86 & 1.26 & 3.30 & 3.61 & 1.53 & 3.54 & 2.52 & 1.07 & 3.34 & 3.32 & 1.00 & 3.35 & 3.85 & 3.67 & 5.29 & 4.21 & 1.21 & 3.41 & 3.92 & 1.01 & 4.15 & 6.24 & 1.40 & 3.48 & 3.68 & 2.85 \\
\bottomrule
\end{tabular}
\end{adjustbox}
\end{table*}

As shown in Table~\ref{tab:u15_weedit_results}, SenseNova-U1.5 substantially outperforms other open-source models and remains competitive with leading proprietary systems. Its strong instruction adherence, text clarity, and background preservation demonstrate robust text-centric editing, where requested textual changes must be accurately executed while unrelated visual content remains intact. The results further suggest improved coordination between semantic editing and visual reconstruction. Translation-related and reasoning-intensive edits, however, remain relatively challenging, indicating room for further improvement in tasks requiring deeper linguistic and contextual understanding.

\paragraph{OmniRef-Bench.}
\begin{table*}[t]
\centering
\caption{\textbf{Quantitative evaluation results on OmniRef-Bench.} We adopt both objective and MLLM-based evaluation protocols.
}
\label{tab:u15_omniref_results}
\renewcommand{\arraystretch}{1}
\setlength{\tabcolsep}{2.5pt}
\begin{adjustbox}{width=\textwidth}
\begin{tabular}{l|cccccc|ccccccc}
\toprule
\multirow{2}{*}{\textbf{Model}} & \multicolumn{6}{c|}{\textbf{Objective Metrics}} & \multicolumn{7}{c}{\textbf{MLLM Evaluation}} \\
\cmidrule(lr){2-7}\cmidrule(lr){8-14}
& \textbf{Subject} & \textbf{Style} & \textbf{Background} & \textbf{Lighting} & \textbf{Pose} & \textbf{Avg.}$\uparrow$ & \textbf{Subject} & \textbf{Background} & \textbf{Style} & \textbf{Lighting} & \textbf{Aesthetics} & \textbf{Instruction} & \textbf{Avg.}$\uparrow$ \\
\midrule
\multicolumn{14}{c}{\textit{Closed-source Models}} \\
\midrule
Nano-Banana-Pro~\cite{deepmind_gemini3proimage_2025} & 0.64 & 0.46 & 0.89 & 0.77 & 0.72 & 0.70 & 8.50 & 8.61 & 8.37 & 8.75 & 8.37 & 8.39 & 8.50 \\
Seedream 4.5~\cite{seedream45} & 0.65 & 0.34 & 0.89 & 0.74 & 0.74 & 0.67 & 8.77 & 8.92 & 6.27 & 8.78 & 8.08 & 7.97 & 8.13 \\
\midrule
\multicolumn{14}{c}{\textit{Open-source Models}} \\
\midrule
\textbf{SenseNova-U1.5 (w/ PE)} & 0.63 & 0.47 & 0.88 & 0.70 & 0.70 & \textbf{0.68} & 8.45 & 8.37 & 7.89 & 8.31 & 7.98 & 7.90 & \textbf{8.15} \\
FLUX.2 [klein]~\cite{flux-2-2025} & 0.63 & 0.21 & 0.87 & 0.69 & 0.74 & 0.63 & 8.25 & 6.32 & 2.18 & 7.69 & 7.95 & 6.53 & 6.49 \\
DreamOmni2~\cite{xia2025dreamomni2} & 0.57 & 0.28 & 0.85 & 0.71 & 0.67 & 0.62 & 6.75 & 3.43 & 3.45 & 6.07 & 6.65 & 4.71 & 5.18 \\
Qwen-Image-Edit-2511~\cite{wu2025qwenimagetechnicalreport} & 0.55 & 0.15 & 0.84 & 0.71 & 0.71 & 0.59 & 6.50 & 4.35 & 1.61 & 6.81 & 6.09 & 4.46 & 4.97 \\
BAGEL~\cite{deng2025bagel} & 0.58 & 0.15 & 0.85 & 0.67 & 0.69 & 0.59 & 6.21 & 3.66 & 1.13 & 6.31 & 5.98 & 4.54 & 4.64 \\
OmniGen2~\cite{wu2025omnigen2} & 0.59 & 0.11 & 0.85 & 0.65 & 0.65 & 0.57 & 4.47 & 2.31 & 0.37 & 4.68 & 6.73 & 4.16 & 3.79 \\
\bottomrule
\end{tabular}
\end{adjustbox}
\end{table*}

In Table~\ref{tab:u15_omniref_results}, SenseNova-U1.5 achieves leading open-source performance on OmniRef-Bench under both objective and MLLM-based evaluation protocols. The gains are particularly pronounced in style consistency and background preservation, indicating that the model can more effectively exploit reference information while minimizing unintended modifications to the target image. Besides, strong subject and pose consistency show that identity, appearance, and structural cues are reliably preserved across diverse editing conditions. These results suggest that U1.5 not only improves reference fidelity, but also achieves a better balance among reference utilization, transformation accuracy, and preservation of the original scene, which is critical for robust multi-reference and customization-oriented editing.

\begin{table}[t]
\centering
\caption{\textbf{Quantitative evaluation results on RISEBench.}
The parameters of the generation component are denoted as \textit{\# Params}. Note that CoT denotes chain-of-thought prompting before outputting the resulting image.}
\label{tab:u15_rise_results}
\begin{adjustbox}{width=0.95\textwidth}
\begin{tabular}{lc|cccc|c|ccc}
\toprule
\textbf{Model} & \textbf{\# Params} & \textbf{Temporal} & \textbf{Causal} & \textbf{Spatial} & \textbf{Logical} & \textbf{Overall}$\uparrow$ & \textbf{IR} & \textbf{AC} & \textbf{VP} \\
\midrule
\multicolumn{10}{c}{\textit{Closed-source Models}} \\
\midrule
GPT-Image-1.5~\cite{GPT-Image-1.5} & - & 54.1 & 60.0 & 62.0 & 21.2 & 50.0 & 69.7 & 92.5 & 94.9 \\
Nano-Banana-Pro~\cite{deepmind_gemini3proimage_2025} & - & 41.2 & 61.1 & 48.0 & 37.6 & 47.2 & 77.0 & 85.5 & 94.4 \\
Nano-Banana~\cite{google2025gemini25flashmodelcard} & - & 25.9 & 47.8 & 37.0 & 18.8 & 32.8 & 61.2 & 86.0 & 91.3 \\
GPT-Image-1~\cite{GPT-Image-1} & - & 34.1 & 32.2 & 37.0 & 10.6 & 28.9 & 62.8 & 80.2 & 94.9 \\
Seedream 4.0~\cite{seedream2025seedream} & - & 12.9 & 12.2 & 11.0 & 7.1 & 10.8 & 58.9 & 67.4 & 91.2 \\
\midrule
\multicolumn{10}{c}{\textit{Open-source Models}} \\
\midrule
\textbf{SenseNova-U1.5 (w/ CoT)} & \textbf{8B} & 38.8 & 52.2 & 42.0 & 20.0 & \textbf{38.6} & 64.7 & 86.8 & 94.8 \\
\textbf{SenseNova-U1.5} & \textbf{8B} & 34.1 & 37.8 & 49.0 & 10.6 & \textbf{33.6} & 59.7 & 85.3 & 95.2 \\
SenseNova-U1-8B-MoT-SFT (w/ CoT)~\cite{diao2026sensenovau1} & 8B & 31.8 & 33.3 & 27.0 & 15.3 & 26.9 & 60.8 & 86.6 & 88.2 \\
SenseNova-U1-8B-MoT-SFT~\cite{diao2026sensenovau1} & 8B & 22.4 & 33.3 & 27.0 & 11.8 & 23.9 & 58.2 & 84.1 & 82.4 \\
Qwen-Image-Edit-2511~\cite{wu2025qwenimagetechnicalreport} & 20B & 21.2 & 18.9 & 31.0 & 4.7 & 19.4 & 49.9 & 71.0 & 91.5 \\
BAGEL (w/ CoT)~\cite{deng2025bagel} & 7B & 5.9 & 17.8 & 21.0 & 1.2 & 11.9 & 45.9 & 73.8 & 80.1 \\
Qwen-Image-Edit-2509~\cite{wu2025qwenimagetechnicalreport} & 20B & 4.7 & 10.0 & 17.0 & 2.4 & 8.9 & 37.2 & 66.4 & 86.9 \\
BAGEL~\cite{deng2025bagel} & 7B & 2.4 & 5.6 & 14.0 & 1.2 & 6.1 & 36.5 & 53.5 & 73.0 \\
FLUX.1-Kontext-Dev~\cite{labs2025flux1kontextflowmatching} & 12B & 2.3 & 5.5 & 13.0 & 1.2 & 5.8 & 26.0 & 71.6 & 85.2 \\
Lumina-DiMOO~\cite{xin2025lumina} & 8B & 2.4 & 1.1 & 4.0 & 1.2 & 2.2 & 34.0 & 50.7 & 72.3 \\
Step1X-Edit~\cite{liu2025step1x} & 12B & 0.0 & 2.2 & 2.0 & 3.5 & 1.9 & 25.1 & 41.5 & 73.5 \\
OmniGen~\cite{xiao2024omnigen} & 3.8B & 1.2 & 1.0 & 0.0 & 1.2 & 0.8 & 22.0 & 32.6 & 55.3 \\
Emu2~\cite{emu2} & 37B & 1.2 & 1.1 & 0.0 & 0.0 & 0.5 & 22.6 & 38.2 & 78.3 \\
\bottomrule
\end{tabular}
\end{adjustbox}
\end{table}

\begin{table*}[!h]
\centering
\caption{\textbf{Quantitative evaluation results on OpenING}.
The parameters of the generation component are denoted as \textit{\# Params}.}
\label{tab:opening_results}
\renewcommand{\arraystretch}{1}
\setlength{\tabcolsep}{2pt}
\begin{adjustbox}{width=\textwidth}
\begin{tabular}{lc|ccccccc|c}
\toprule
\textbf{Model} & \textbf{\# Params} & \textbf{Complete} & \textbf{Quality} & \textbf{Richness} & \textbf{Correct} & \textbf{Human Align.} & \textbf{IT Coherency} & \textbf{Multi-step} & \textbf{Overall}$\uparrow$ \\
\midrule
\multicolumn{10}{c}{\textit{Closed-source Models}} \\
\midrule
Nano-Banana~\cite{google2025gemini25flashmodelcard} & - & 9.34 & 8.58 & 8.00 & 9.17 & 8.88 & 9.27 & 8.70 & 8.85 \\
Wan-Weaver~\cite{xing2026wan} & - & 9.41 & 8.32 & 8.03 & 8.90 & 8.69 & 8.78 & 8.56 & 8.67 \\
GPT-4o~\cite{hurst2024gpt}+DALL-E3~\cite{betker2023dalle3} & - & 8.66 & 8.01 & 7.42 & 7.98 & 8.77 & 8.15 & 8.38 & 8.20 \\
Gemini~\cite{team2023gemini}+Flux~\cite{flux2024} & - & 7.58 & 7.26 & 6.48 & 7.03 & 7.98 & 6.98 & 7.33 & 7.23 \\
\midrule
\multicolumn{10}{c}{\textit{Open-source Models}} \\
\midrule
\textbf{SenseNova-U1.5 (w/ CoT)} & \textbf{8B} & 9.13 & 9.21 & 8.51 & 9.19 & 9.45 & 9.58 & 9.21 & \textbf{9.18} \\
SenseNova-U1-SFT (w/ CoT)~\cite{diao2026sensenovau1} & 8B & 9.14 & 9.03 & 8.43 & 9.08 & 9.35 & 9.40 & 9.09 & 9.07 \\
SEED-X~\cite{ge2024seed} & 17B & 5.65 & 6.07 & 4.92 & 5.77 & 7.03 & 5.72 & 5.72 & 5.84 \\
Emu3~\cite{wang2024emu3} & 8B & 5.90 & 5.96 & 5.52 & 5.43 & 6.47 & 5.66 & 5.37 & 5.76 \\
Anole~\cite{chern2024anole} & 7B & 6.27 & 6.02 & 5.28 & 5.06 & 6.91 & 4.90 & 5.81 & 5.75 \\
SEED-LLaMA~\cite{ge2023making} & 14B & 5.59 & 5.50 & 4.61 & 4.59 & 6.50 & 4.43 & 5.13 & 5.19 \\
VILA-U~\cite{wu2024vila} & 7B & 5.60 & 5.14 & 4.68 & 4.78 & 5.69 & 4.74 & 4.79 & 5.06 \\
Show-o~\cite{xie2024show} & 1.3B & 4.37 & 4.79 & 3.83 & 3.76 & 5.78 & 4.04 & 4.33 & 4.41 \\
MiniGPT-5~\cite{zheng2023minigpt} & 0.86B & 3.91 & 4.50 & 3.61 & 3.63 & 5.51 & 3.56 & 4.10 & 4.12 \\
NExT-GPT~\cite{wunext} & 1.3B & 3.89 & 4.25 & 3.35 & 3.61 & 5.35 & 3.32 & 3.85 & 3.95 \\

\bottomrule
\end{tabular}
\end{adjustbox}
\end{table*}

\textbf{Reasoning-centric Editing.}
Beyond direct instruction following, we further evaluate edits that require the model to infer temporal, causal, spatial, or logical consequences before modifying the image, using RISEBench~\cite{zhao2025envisioning}.

\paragraph{RISEBench.}

As shown in Table~\ref{tab:u15_rise_results}, SenseNova-U1.5 already outperforms the evaluated open-source baselines without CoT, while explicit reasoning provides a further substantial improvement. The gains are most pronounced for causal, logical, and temporal edits, indicating that CoT is particularly effective when the required transformation must first be inferred rather than directly localized. In contrast, spatial editing benefits less consistently, suggesting that explicit reasoning is most valuable for implicit multi-step transformations rather than directly grounded edits.

\begin{table*}[t]
\centering
\caption{\textbf{Quantitative evaluation results on VBVR-Pro-Bench.}
We evaluate two model settings: image generation and interleaved generation. We report both in-domain (ID) and out-of-domain (OOD) scores as percentages, together with category-wise performance.}
\label{tab:vbvr_pro_bench}
\renewcommand{\arraystretch}{1}
\setlength{\tabcolsep}{4pt}
\begin{adjustbox}{width=\textwidth}
\begin{tabular}{l|c|cccccc|cccccc}
\toprule
\multirow{2}{*}{\textbf{Model}} & \multirow{2}{*}{\textbf{Overall}$\uparrow$}
& \multicolumn{6}{c|}{\textbf{In-Domain by Category}}
& \multicolumn{6}{c}{\textbf{Out-of-Domain by Category}} \\
\cmidrule(lr){3-8}\cmidrule(lr){9-14}
& & \textbf{Avg.} & \textbf{Abst.} & \textbf{Know.} & \textbf{Perc.} & \textbf{Spat.} & \textbf{Trans.}
& \textbf{Avg.} & \textbf{Abst.} & \textbf{Know.} & \textbf{Perc.} & \textbf{Spat.} & \textbf{Trans.} \\
\midrule
\multicolumn{14}{c}{\textit{Closed-source Models (Image Generation)}} \\
\midrule
Seedream-5.0-Pro~\cite{Seedream5} & 55.7 & 48.5 & 67.3 & 34.9 & 67.9 & 44.1 & 30.7 & 62.9 & 68.9 & 56.5 & 74.4 & 55.9 & 22.3 \\
Qwen-Image-2.0~\cite{zhao2026qwenimage20technicalreport} & 31.3 & 24.8 & 35.0 & 22.0 & 30.0 & 18.7 & 18.6 & 37.8 & 46.3 & 29.2 & 44.0 & 38.4 & 8.8 \\
\midrule
\multicolumn{14}{c}{\textit{Open-source Models (Image Generation)}} \\
\midrule
FLUX.2-dev~\cite{flux-2-2025} & 15.7 & 10.8 & 11.4 & 12.2 & 9.6 & 11.0 & 9.3 & 20.6 & 26.8 & 20.5 & 20.7 & 24.1 & 8.5 \\
Qwen-Image-Edit~\cite{wu2025qwenimagetechnicalreport} & 13.4 & 10.8 & 12.0 & 9.2 & 13.4 & 12.0 & 7.9 & 15.9 & 23.9 & 7.8 & 15.8 & 18.2 & 9.1 \\
BAGEL ~\cite{deng2025bagel} & 8.9 & 6.6 & 5.0 & 9.5 & 8.9 & 5.0 & 3.8 & 11.1 & 27.2 & 3.9 & 8.2 & 2.8 & 13.3 \\
\midrule
\multicolumn{14}{c}{\textit{Closed-source Models (Interleaved Generation)}} \\
\midrule
Nano-Banana-Pro~\cite{deepmind_gemini3proimage_2025} & 56.4 & 48.0 & 67.4 & 47.3 & 68.2 & 31.3 & 24.6 & 64.8 & 75.1 & 61.9 & 73.9 & 58.5 & 24.3 \\
GPT-Image-2~\cite{gpt_image_2} & 50.7 & 42.8 & 59.3 & 35.6 & 57.1 & 22.7 & 42.4 & 58.7 & 54.0 & 51.3 & 71.2 & 48.0 & 33.5 \\
\midrule
\multicolumn{14}{c}{\textit{Open-source Models (Interleaved Generation)}} \\
\midrule
\textbf{SenseNova-U1.5} & \textbf{68.2} & \textbf{67.6} & 78.4 & 75.2 & 70.5 & 60.8 & 48.7 & \textbf{68.9} & 69.8 & 59.9 & 71.6 & 67.8 & 65.0 \\
VBVR-SenseNova-U1~\cite{xu2026vbvrproscalableverifiablesuite} & 40.8 & 46.9 & 46.3 & 35.1 & 49.7 & 42.5 & 67.4 & 34.7 & 39.5 & 39.3 & 30.9 & 48.0 & 26.3 \\
ThinkMorph ~\cite{gu2025thinkmorph} & 15.4 & 11.3 & 13.0 & 9.2 & 13.4 & 16.2 & 4.4 & 19.5 & 23.8 & 20.6 & 18.3 & 25.3 & 11.4 \\
\bottomrule
\end{tabular}
\end{adjustbox}
\end{table*}

\begin{table*}[t]
\centering
\caption{\textbf{Quantitative evaluation results on Uni-MMMU-GaU and RealUnify-GEU.}
For Uni-MMMU-GaU, we report text accuracy (T) for all tasks and sample-level accuracy for the multi-step Maze and Sliding Puzzle tasks. For RealUnify-GEU, MR, MT, AF, and CN denote mental reconstruction, mental tracking, attentional focusing, and cognitive navigation, respectively.}
\label{tab:unified_reasoning_results}
\renewcommand{\arraystretch}{1}
\setlength{\tabcolsep}{5pt}
\begin{adjustbox}{max width=\textwidth}
\begin{tabular}{lc|ccccc|ccccc}
\toprule
\multirow{2}{*}{\textbf{Model}} & \multirow{2}{*}{\textbf{\# Params}} & \multicolumn{5}{c|}{\textbf{Uni-MMMU-GaU}} & \multicolumn{5}{c}{\textbf{RealUnify-GEU}} \\
\cmidrule(lr){3-7}\cmidrule(lr){8-12}
& & \textbf{Jigsaw-T} & \textbf{Maze-T} & \textbf{Sliding-T} & \textbf{Geometry-T} & \textbf{Avg}$\uparrow$ & \textbf{MR} & \textbf{MT} & \textbf{AF} & \textbf{CN} & \textbf{Avg}$\uparrow$ \\
\midrule
\textbf{SenseNova-U1.5} & \textbf{8B} & 86.0 & 17.3 & 0.0 & 15.0 & \textbf{29.6} & 38.0 & 84.0 & 51.0 & 52.0 & \textbf{56.3} \\
SenseNova-U1-SFT~\cite{diao2026sensenovau1} & 8B & 87.3 & 28.6 & 0.0 & 24.2 & 35.0 & 36.0 & 63.0 & 51.0 & 40.0 & 47.5 \\
BAGEL~\cite{deng2025bagel} & 7B & 48.0 & 0.0 & 1.2 & 32.8 & 20.5 & 38.0 & 25.0 & 52.0 & 28.0 & 35.8 \\
OmniGen2~\cite{wu2025omnigen2} & 4B & 48.0 & 0.0 & 0.0 & 5.7 & 13.4 & 42.0 & 24.0 & 38.0 & 19.0 & 30.8 \\
Ovis-U1~\cite{wang2025ovis} & 1.2B & 53.0 & 0.0 & 0.0 & 3.5 & 14.1 & 38.0 & 25.0 & 31.0 & 24.0 & 29.5 \\
\bottomrule
\end{tabular}
\end{adjustbox}
\end{table*}

\subsection{Interleaved Generation}  

\textbf{Interleaved Generation.}
We evaluate this capability on OpenING~\cite{zhou2025opening} and VBVR-Pro-Bench~\cite{xu2026vbvrproscalableverifiablesuite} that integrate understanding, reasoning, and generation for open-ended interleaved generation and generation-mediated visual reasoning.

\paragraph{OpenING.}
In Table~\ref{tab:opening_results}, SenseNova-U1.5 with CoT achieves leading overall performance on OpenING, improving over SenseNova-U1 and outperforming the compared proprietary pipelines. Strong results in image-text coherency, human alignment, and multi-step consistency demonstrate reliable open-ended interleaved generation.

\paragraph{VBVR-Pro-Bench.}
In Table~\ref{tab:vbvr_pro_bench}, SenseNova-U1.5 achieves new state-of-the-art results across cognitive faculties, showing its ability to reason directly through interleaved text-image generation. It also generalizes strongly to out-of-domain tasks, outperforming leading proprietary models such as Nano-Banana-Pro~\cite{deepmind_gemini3proimage_2025} and GPT-Image-2~\cite{gpt_image_2}.

\textbf{Unified Reasoning.}
Uni-MMMU~\cite{zou2025unimmmu} evaluates generation-aided understanding (GaU), while RealUnify~\cite{shi2025realunify} evaluates generation-enhanced understanding (GEU) across reconstruction, tracking, focusing, and navigation.

\paragraph{Uni-MMMU-GaU and RealUnify-GEU.}
In Table~\ref{tab:unified_reasoning_results}, SenseNova-U1.5 achieves leading performance on RealUnify-GEU with a considerable improvement over SenseNova-U1-SFT, while remaining competitive with the previous SenseNova model on Uni-MMMU-GaU. These findings show that generation in SenseNova-U1.5 is not only an output modality, but can also function as an intermediate representation that supports multimodal reasoning and understanding.

\section{Conclusion}

We present SenseNova-U1.5 as a further step toward native multimodal systems in which visual understanding and generation are learned within a common computational substrate rather than connected through separate perceptual and generative pipelines. Our extensive results demonstrate that this direction can scale beyond basic image synthesis: a compact native representation can simultaneously sustain strong understanding while impressively supporting high-fidelity generation, editing, interleaved generation, and reasoning-intensive visual tasks.

More broadly, SenseNova-U1.5 suggests that the central opportunity of multimodal unification is not architectural simplification, but capability interaction. Reasoning can improve how visual content is created and transformed, while generation can itself become part of the model’s internal process for solving visual problems. This blurs the conventional boundary between perception and creation, and points toward a class of foundation models in which seeing, reasoning, and generating are no longer trained as loosely coupled functions, but emerge as different expressions of a shared visual intelligence. We view this as an important step toward multimodal models that do not merely combine modalities, but develop a unified mechanism for interpreting, imagining, and acting upon visual worlds.

\section{Contributors}

The list is organized by contribution role, with individuals listed alphabetically by their first name within each category.\\

\textbf{Project Sponsor and Advisor}:\\
Dahua Lin

\textbf{Senior Project Lead}: \\
Lei Yang, Lewei Lu, Quan Wang, Ruihao Gong, Wenxiu Sun, Ziwei Liu

\textbf{Project Lead}: \\
Haiwen Diao, Jiahao Wang

\textbf{Core Contributor}: \\
Chenjing Ding, Hanming Deng, Jiangnan Chen, Ruixi Zhang, Ruohui Wang, Wenwen Tong, Xiangyu Fan, Yubo Wang, Yue Zhu, Yuwei Niu, Zhengqi Bai, Zhiqian Lin, Zhitao Yang, Zhongang Cai

\textbf{Contributor}: \\
Bo Yang, Chen Feng, Chengguang Lv, Guangjia Liu, Guanlin Wang, Hanyu Zhang, Haojia Yu, Hongcan Xiao, Hongli Wang, Huan Wu, Huaping Zhong, Jian Fang, Jianan Fan, Jiaqi Li, Jiefan Lu, Jing Zuo, Jingcheng Ni, Junxiang Xu, Linjun Dai, Mutian Xu, Peishen Yan, Penghao Wu, Ruijie Mao, Ruisi Wang, Shihao Bai, Shuang Yang, Shuya Yang, Shuyan Zheng, Silei Wu, Siying Li, Tao Chu, Tianbo Zhong, Tongxi Zhou, Weichao Luo, Weichen Fan, Wenhao Jia, Wenjie Gao, Xiangli Kong, Yan Li, Yang Yong, Zimo Wen, Zixuan Qian

\textbf{Acknowledgement}: \\
We would like to thank Boyu Guan, Chen Wei, Chenyang Gu, Fanzhou Wang, Haoge Deng, Houyuan Chen, Huchuan Lu, Jiaxu Li, Kaizhe Yang, Lianqiang Shi, Oscar Qian, Qingping Sun, Wanqi Yin, Wenjie Ye, Xiaotong Li, Xuehai Bai, Yijia Fan, Yue Huang, Yukun Wei, Yves Wang, and Zongpu Zhang for their valuable support for this project, including data preparation, model evaluation, infrastructure support, architecture analysis, and helpful discussions.

\clearpage

\bibliographystyle{plainnat}
\bibliography{main}

\clearpage

\end{document}